\documentclass[10pt,twocolumn,letterpaper]{article}

\usepackage{cvpr}
\usepackage{times}
\usepackage{epsfig}
\usepackage{graphicx}
\usepackage{amsmath}
\usepackage{amssymb}
\usepackage{booktabs}
\usepackage{multirow}
\usepackage{subfigure}

\usepackage[pagebackref=true,breaklinks=true,letterpaper=true,colorlinks,bookmarks=false]{hyperref}

\cvprfinalcopy 

\def\cvprPaperID{4584} 

\ifcvprfinal\fi
\begin{document}

\title{Appending Adversarial Frames for Universal Video Attack}

\author{
Zhikai Chen\\
Xi'an Jiaotong University\\
{\tt\small zhikai\_chen@outlook.com}
\and
Lingxi Xie\\
Huawei Noah's Ark Lab\\
{\tt\small 198808xc@gmail.com}
\and
Shanmin Pang\\
Xi'an Jiaotong University\\
{\tt\small pangsm@xjtu.edu.cn}
\and
Yong He\\
Xi'an Jiaotong University\\
{\tt\small hy0275@stu.xjtu.edu.cn}
\and
Qi Tian\\
Huawei Noah's Ark Lab\\
{\tt\small tian.qi1@huawei.com}
}

\maketitle

\begin{abstract}
    There have been many efforts in attacking image classification models with adversarial perturbations, but the same topic on video classification has not yet been thoroughly studied. This paper presents a novel idea of video-based attack, which appends a few dummy frames (e.g., containing the texts of `thanks for watching') to a video clip and then adds adversarial perturbations \textbf{only} on these new frames. Our approach enjoys three major benefits, namely, a high success rate, a low perceptibility, and a strong ability in transferring across different networks. These benefits mostly come from the common dummy frame which pushes all samples towards the boundary of classification. On the other hand, such attacks are easily to be concealed since most people would not notice the abnormality behind the perturbed video clips. We perform experiments on two popular datasets with six state-of-the-art video classification models, and demonstrate the effectiveness of our approach in the scenario of universal video attacks.
\end{abstract}


\section{Introduction}
Deep neural networks, while being powerful in learning from complicated visual data, are vulnerable to small noise known as adversarial perturbations. Researchers designed a lot of attacking algorithms to add imperceptible perturbations onto well-trained neural networks so that the prediction is dramatically destroyed, and successful scenarios include image classification~\cite{kurakin2016adversarial,Moosavi_Dezfooli_2016,aleks2017deep}, object detection and segmentation~\cite{Xie_2017}, super-resolution~\cite{Choi_2019_ICCV}, visual question answering~\cite{xu2017fooling}, image captioning~\cite{Choi_2019_ICCV}, \textit{etc}. Researchers conjectured that adversaries are closely related to the working mechanism as well as explainability of deep neural networks~\cite{szegedy2013intriguing,goodfellow2014explaining}, and both adversarial attack and defense have been attracting a lot of attentions in both academia and industry.

While adversarial attack and defense have been covered in a wide range of vision tasks on still images, the same topics on video data have not been carefully studied. Existing methods for video-based adversarial attacks mostly involved adding perturbations to a few frames in the target video clips, which ignored the fact that images and videos have different properties. In particular, video is a sequence of images on which neighboring frames are closely correlated. Adding individual noise to each frame can increase the perceptibility of the attack. On the other hand, people are less sensitive to some specific types of contents (\textit{e.g.}, ending frames), which offers extra opportunities for designing effective attacking algorithms.

\begin{figure}[!t]
\centering\includegraphics[width=0.48\textwidth]{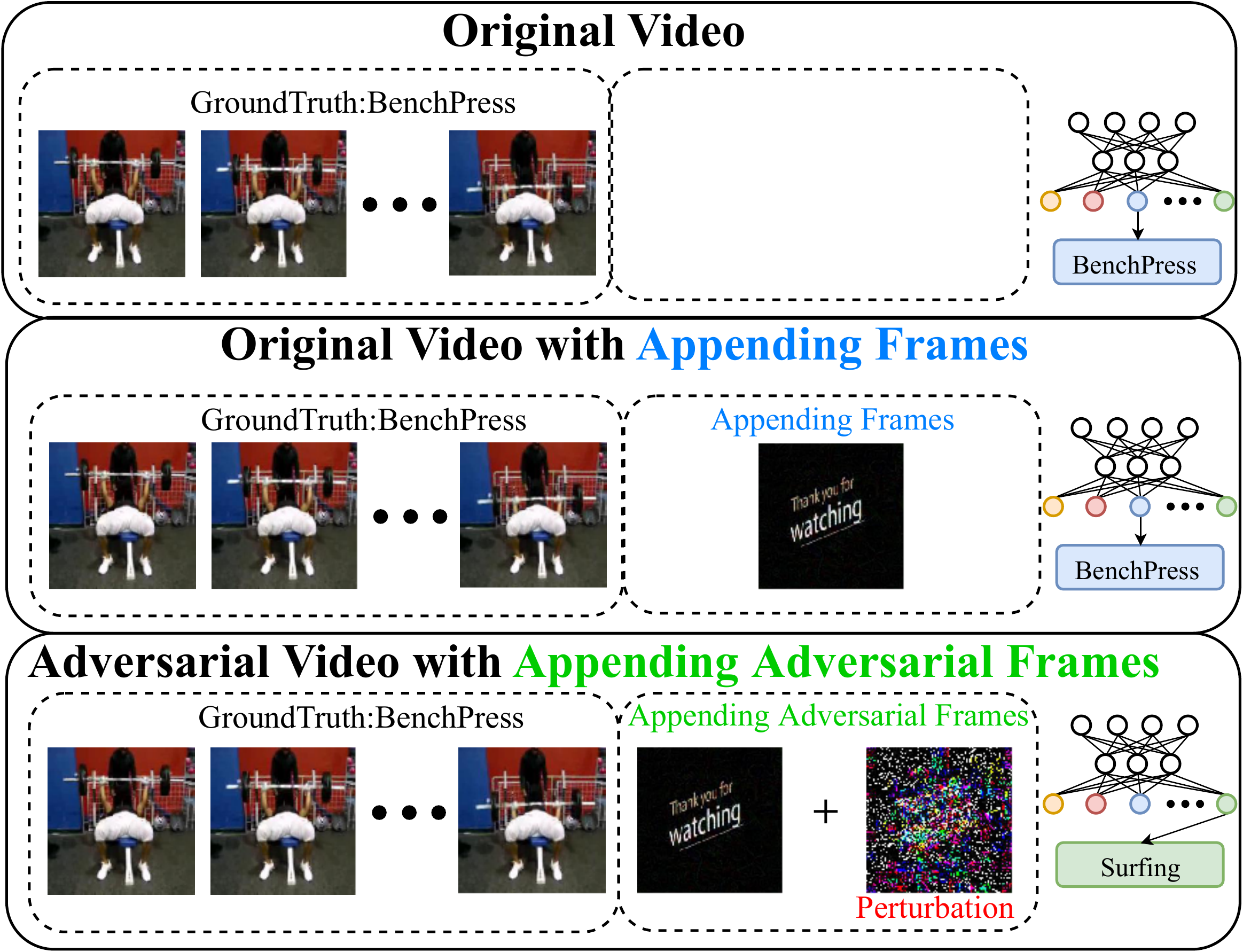}
\vspace{-.2cm}
\caption{An example of appending adversarial frames on video data.  One can see that the label of the original video is `BenchPress'.
When adding an appedning frame with the texts of `thanks for watching' to the video, it does not fool the recognition model, and the label of the generated new video is still `BenchPress'. 
However, when continuing adding the computed perturbation to the appending frame, the recognition model corruptly predicts the final adversarial video as `Surfing'.}
\label{fig:illustration}
\vspace{-.2cm}
\end{figure}

In this paper, we follow this observation and propose a novel approach for video-based adversarial attack. The idea is simple: instead of adding perturbations to a normal video frame, we append a few dummy frames (\textit{e.g.}, all-black images with the texts of `thanks for watching' on it, see Figure~\ref{fig:illustration}). This operation, while having a very large perceptibility (calculated by accumulating pixel-wise differences), can be easily concealed because most people would believe them to be the common frames to conclude the video. In this manner, we provide an alternative starting point (target input) of attack. In what follows, perturbations are only allowed on these new frames, \textit{i.e.}, all the original frames remain clean. We also consider a simple defender to our attack which detects significant feature differences between neighboring frames, and hide from it by eliminating such quantities in constructing the perturbations.

Compared to a regular attack, our approach enjoys three-fold benefits. \textbf{First}, a high success rate. Note that the dummy frames are shared among all videos, which weakens the extent of the videos being discriminative. In other words, this operation pushes each video (as a point in the feature space) towards the boundary of classification, and thus adversarial attack becomes much easier. \textbf{Second}, a low perceptibility. For the same reason, the cost required to successfully attack each video is made much lower. Although the perturbations are still added to individual frames, the chance that they are detected by human sense is smaller. \textbf{Third}, a strong ability in transferring across different networks. For the same reason, it is more likely that the perturbations that make a video clip fail in a specific network can also make it fail on other networks. In other words, our approach has a higher success rate in black-box attacks.

We evaluate our approach on two popular video classification datasets, namely, UCF101~\cite{soomro2012ucf101} and HMDB51~\cite{kuehne2011hmdb}. We select six victim models from a wide range of video classification methods, including CNN+LSTM~\cite{donahue2014longterm}, C3D~\cite{Tran_2015}, ResNet3D~\cite{Hara_2018}, P3D~\cite{qiu2017learning}, and I3D~\cite{carreira2017quo} on both ResNet~\cite{He_2016_CVPR} and Inception~\cite{Szegedy_2016}. We start with the white-box attacks, in which the parameters of the victim models are known to the attacker, and then transfer the perturbations computed on each video to other unseen models, \textit{i.e.}, the black-box attacks. Our approach enjoys superior success rate while the perceptibility added to the dummy frames is much smaller than that added by the baseline approaches to the original video frames. Therefore, we demonstrate the effectiveness of our approach in \textbf{universal} video attacks.

The main contributions of this paper can be summarized as fellows:

\begin{itemize}
\vspace{-0.20cm}
\setlength{\itemsep}{-0.2ex} 
\item We propose to append adversarial frames to videos, which is an effective attacking method with strong transferability while being easily concealed.
\item We design an intermediate layer representation-like attack method to further decrease the perceptibility of adversaries to a simple defender in feature space.
\item We investigate the attack performance by only attacking a certain spatial part of frames.
\item We provide an explanation of the mechanism of our attacking method, which suggests that the diversity of perturbation can actually be constrained by appending adversarial frames.
\end{itemize}

\section{Related Work}

\textbf{Video Recognition Models.} 
The power of deep learning architectures is not only shown in image classification (ImageNet~\cite{Deng_2009}), but also shown in video recognition.
There are many successful video recognition models. For instance, \cite{donahue2014longterm} proposed a class of recurrent long-term models that can be jointly trained to learn temporal dynamics and convolutional perceptual representations, and demonstrated superior performance on recognition and description of images and videos. \cite{Tran_2015} addressed the problem of learning spatio tempora features for videos using 3D ConvNets. Noting that the training of 3D ConvNets is computationally extensive, \cite{carreira2017quo} introduced a two-stream Inflated 3D (I3D) ConvNets, which built upon on 2D kernels but inflated filters and pooling kernels into 3D. 
Another representative strategy to reduce training cost was proposed in \cite{qiu2017learning}, which simulated 3D convolutions with 2D convolutions on spatial domain plus 1D convolutions on adjacent feature maps in time.
Recently,~\cite{Hara_2018} examined the architectures of various 3D CNNs from relatively shallow networks to very deep ones on current video datasets. 

\textbf{Adversarial Image Attack.} 
There are fruitful attack methods in the literature. Among the first to introduce adversarial examples against deep neural networks was \cite{szegedy2013intriguing}. After that, Goodfellow \textit{et al.} \cite{goodfellow2014explaining} used the sign of the gradient to propose a fast attack method called Fast Gradient Sign Method (FGSM). FGSM seeks the direction that can maximize the classification errors to update each pixel. 
The subsequent method I-FGSM~\cite{kurakin2016adversarial} extends FGSM by more iterations, and can generate adversarial examples in the physical world.
In \cite{aleks2017deep}, an iterative method called Projection Gradient Descent (PGD) was proposed.  
PGD makes the perturbations project back to the $\epsilon$-ball which center is the original data when perturbations over the $\epsilon$-ball. \cite{Moosavi_Dezfooli_2016} proposed \textit{DeepFool} to find the closest distance from the original input to the decision boundary of adversarial examples.
\cite{moosavi2017universal} analyzed universal perturbations and its relationship between different classification regions of decision boundary. \cite{liu2016delving} studied the transferability of both non-targeted and targeted adversarial examples, and proposed an ensemble-based approaches to generate adversarial examples with stronger transferability. \cite{sabour2015adversarial} performed a targeted attack by minimizing the distance of the representation of intermediate neural network layers instead of the output layer. There also exists other methods for white-box attack, \textit{e.g.} Jacobian-based Saliency Map Attack (JSMA)~\cite{Papernot_2016} and Elastic net attack (EAD)~\cite{chen2017ead}. Except the white-box image attack, some black-box attack methods \textit{e.g.} Zeroth-order optimization attack (ZOO)~\cite{Chen2017ZOOZO} used a black-box method to estimate the adversarial gradient. HopSkipJumpAttack~\cite{chen2019hopskipjumpattack},  which estimated the gradient direction using binary information at the decision boundary, is also a family of black-box algorithms. 

\textbf{Adversarial Video Attack.}
There are several attack methods proposed for generating video adversarial examples. Wei \textit{et al.}\cite{wei2018sparse} claimed that they were the first to explore the adversarial examples in videos. In their paper, they mainly investigated the sparsity and propagation of adversarial perturbations across video frames. Recently, \cite{Li2018StealthyAP} showed that we can use a Generative Adversarial Networks (GANs) like architecture to generate perturbations in real-time video classifier. \cite{jiang2019blackbox} was the first work on black-box video attacks against video recognition models. In this paper, we use an optimization based method to find adversarial perturbations. 
Being different from previous works that add perturbations to the original videos, we append the adversarial frames to the original videos. To the best of our knowledge, we are the first to propose to append adversarial frames to videos, and we also develop a method to make the appending adversarial frames be similar with the original ones in the intermediate layer feature representation.

\section{Methodology}

\subsection{Problem Definition}
We use $\mathbb{J}(\cdot ; \boldsymbol{\boldsymbol{\theta}})$ to denote the threat model with network parameter $\boldsymbol{\boldsymbol{\theta}}$. We always consider the threat model as a deep neural network, which means the classifier $\mathbb{J}(\cdot ; \boldsymbol{\boldsymbol{\theta}})$ is a complicated yet  differentiable function. $\ell(\cdot , \cdot)$ is a loss function to measure the difference between the prediction and the ground truth label, and in this paper we use well-known cross-entropy as the loss function. 
Given a video $\mathbf{X}$ and its corresponding ground truth label $y$, to harvest the adversarial video $\mathbf{\hat{X}}$ with its prediction label being different from $y$, our goal is to maximize the loss $\ell(\mathbf{1}_y, {J}(\mathbf{\hat{X}};\boldsymbol{\boldsymbol{\theta}}))$ for $\mathbf{\hat{X}}$, where $\mathbf{\hat{X}} = \mathbf{X} + \mathbf{E}$, and $\mathbf{E}$ denotes the adversarial perturbation.
To be concealed, the norm of the perturbation $\mathbf{E}$ is constrained in to be small. $\mathbf{1}_y$ is the one-hot encoding of the ground truth label $y$. 

\subsection{Motivation}
The existed video attack methods can be considered as adding adversarial perturbations to original video frames. In this paper, we define them as \textbf{Basic Attack Methods (BAM)}. 
More specifically, Let $\mathbf{X}=\{f_1, f_2, \dots, f_T\}$ be a video that contains $T$ frames. 
Assume its corresponding perturbations $\mathbf{E} = \{p_1, p_2, \dots, p_T\}$, then the adversarial video can be represented as $\mathbf{\hat{X}} = \{f_1 \oplus p_1, f_2 \oplus p_2, \dots, f_T \oplus p_T\}$, where $\oplus$ denotes pixel-wise addition.

The regular attack methods belonging to this group have the following limitations: 
(i) The operation $\oplus$ needs a high authority to manipulate frames of the video, which is difficult in some situations. 
(ii) BAM is easy to be perceived, because the neighboring frames: $\{f_1, f_2, \dots, f_T\}$ are related while the perturbations: $\{p_1, p_2, \dots, p_T\}$ are not generated in a uniform pattern. 
(iii) Wei \textit{et al.}\cite{wei2018sparse} demonstrated that we can generate adversarial videos using sparse perturbations. 
However, if the goal is to generate a universal perturbation across videos, their method needs a huge amount of adversarial perturbations and the magnitude of $\epsilon$ should be set large, otherwise the fooling rate under the constraint that $||E||_p<\epsilon$ is often low.
(iv) Last but not least, the adversarial perturbations generated by the existed methods have a weak transferability across models with different architecture $\boldsymbol{\boldsymbol{\theta}}_k$. 
All of the limitations are referred in Sec.~\ref{sec:experiments}. Those problems inspire us to find a more effective way to generate perturbations.

\subsection{Appending Adversarial Frames}
In this paper, we propose a method called \textbf{Appending Adversarial Frames Method (A\textsuperscript{2}FM)} to solve the limitations of existed methods. A\textsuperscript{2}FM generates the adversarial videos through appending adversarial frames to the original video. 
The appending strategy needs less authority to apply attack to video data. Because although there are a lot of videos not allowed to modify, they are allowed to append as this operation does not change the content of videos. Furthermore, appending is a more inconspicuous way to attack. 
As described earlier, appending like ``thanks for watching" dummy frames can hardly be concealed as most people would believe them to be the common frames to conclude the video. 
What is more, compared with elaborately designing perturbations for different videos, appending universal adversarial frames is more general to transfer to different types of videos.

Let $\mathbf{X}\in\mathbb{R}^{T \times W \times H \times C}$ be the original video, where $T$ denotes the number of frames, and  $W, H, C$  denote the width, height, and channel of each frame, respectively. Let $\mathbf{\Delta}\in\mathbb{R}^{\Delta T \times W \times H \times C}$ be the adversarial frames without perturbations, and $\mathbf{\hat{\Delta}}$ be its adversarial frames with perturbations. Thus, $\mathbf{E}=\mathbf{\hat{\Delta}}-\mathbf{\Delta}$ is the adversarial perturbations, and correspondingly, $\mathbf{\hat{X}}\in\mathbb{R}^{(T + \Delta T) \times W \times H \times C}$ is its adversarial video.
We use A\textsuperscript{2}FM to find $\mathbf{E}$ to maximize the difference between the output from video classification models which take $\mathbf{\hat{X}}$ and $\mathbf{X}$ as input:

\begin{equation}\label{eq:eq1}
\arg\min_{\mathbf{E}} \lambda ||\mathbf{E}||_{p} -\ell(\mathbf{1}_y, {\mathbb{J}}(\mathbf{\hat{X}};\boldsymbol{\theta}))
\end{equation}
where $||\mathbf{E}||_p$ is the $\ell_p$ norm (we select $p=\infty$ in this paper) of $\mathbf{E}$, which is used to measure the magnitude of the adversarial perturbations. The parameter $\lambda$ is the weight applied to balance different items in the objective function.

If the goal is to misclassify the generated adversarial video  $\mathbf{\hat{X}}$ to a  predefined label (\textit{i.e.}, the target label). then we can modify the problem to minimize the difference between the target label and the prediction.

\begin{equation}\label{eq:eq2}
\arg\min_{\mathbf{E}} \lambda ||\mathbf{E}||_{p} + \ell(\mathbf{1}_{y^*}, {\mathbb{J}}(\mathbf{\hat{X}};\boldsymbol{\theta}))
\end{equation}
where $y^*$ is the targeted label. 

\subsection{Variants of A\textsuperscript{2}FM}
As the method A\textsuperscript{2}FM we mentioned above, we can base on its strengths and properties to develop a series of variants to overcome different problems in various attack scenarios.

\textbf{Appending Adversarial Frames Method Across Videos (\textbf{A\textsuperscript{2}FM-AV}).} Although we can compute different perturbations for different videos to generate a series of specific adversarial frames, as showed in Eq.~\ref{eq:eq1}, it is also possible to find a \textit{video-agnostic} adversarial perturbation, which can apply to any input video for a certain video-classification method. 
\begin{equation}\label{eq:eq3}
\arg\min_{\mathbf{E}} {\lambda ||\mathbf{E}||_{p}} - \sum_{n=1}^N\alpha_n\ell(\mathbf{1}_{y_n}, {\mathbb{J}}(\mathbf{\hat{X}}_n;\boldsymbol{\theta}))
\end{equation}
where $N$ is the total number of videos for finding the universal adversarial perturbation.
The parameter $\alpha_n$ is the contribution of the $n$-th video to generate the adversarial perturbation,  and $\mathbf{\hat{X}}_n$ is the $n$-th adversarial video.  

\textbf{Appending Adversarial Frames Method Across Models (\textbf{A\textsuperscript{2}FM-AM}).} We can also develop a \textit{model-agnostic} attack method to generate a universal adversarial perturbation across models. Specifically, we use an ensemble-based method to solve the problem:
\begin{equation}\label{eq:eq4}
\arg\min_{\mathbf{E}} \lambda ||\mathbf{E}||_{p} - \sum_{k=1}^K\beta_k\ell(\mathbf{1}_{y_{k}}, {\mathbb{J}}(\mathbf{\hat{X}};\boldsymbol{\theta}_k))
\end{equation}
where $K$ is the total number of models, and ${\mathbb{J}(\cdot ; \boldsymbol{\theta}_k)}$ is the $k$-th model.
Similar to $\alpha_n$, $\beta_k$ is the weight of model ${\mathbb{J}(\cdot ; \boldsymbol{\theta}_k)}$.

\textbf{Appending Adversarial Frames Method with Feature Similarity (\textbf{A\textsuperscript{2}FM-FS}).} 
Let $\Delta_s$ denotes a serial of frames selected from the original video randomly, where $\Delta_s$ has the same dimensions compared with $\Delta$. Let $\phi_l$ be the mapping from an image to its internal DNNs representation at layer $l$. To obtain a similar latent feature perturbation, our goal is to find the perturbation $E$ that makes the distance between $\phi_l(\mathbf{\Delta_s})$ ~and ~$\phi_l(\mathbf{\hat{\Delta}})$ as small as possible. 
When $l=0$, we replace $\Delta_s$ with $\Delta$, and therefore in this case we use the pixel level distance to constrain $\mathbf{E}$, that is, ${||\phi_0(\mathbf{\Delta})-\phi_0(\mathbf{\hat{\Delta}})||_p}$ $=$ ${||\mathbf{E}||_p}$. With Feature-Similar constraint to the perturbation, we optimize the following objective function:

\begin{equation}\label{eq:eq5}
\begin{split}
\arg\min_{\mathbf{E}} 
\lambda ||\mathbf{E}||_{p} - \ell(\mathbf{1}_{y}, {\mathbb{J}}(\mathbf{\hat{X}};\boldsymbol{\theta})) \\
+ \lambda_l{||\phi_l(\mathbf{\Delta_s})-\phi_l(\mathbf{\hat{\Delta}})||_p} 
\end{split}
\end{equation}
where $\lambda_l$ is the regularization parameter of the layer $l$.

\subsection{Relationship between Different Attacks}
\begin{figure}[!t]
\centering\includegraphics[width=0.45\textwidth]{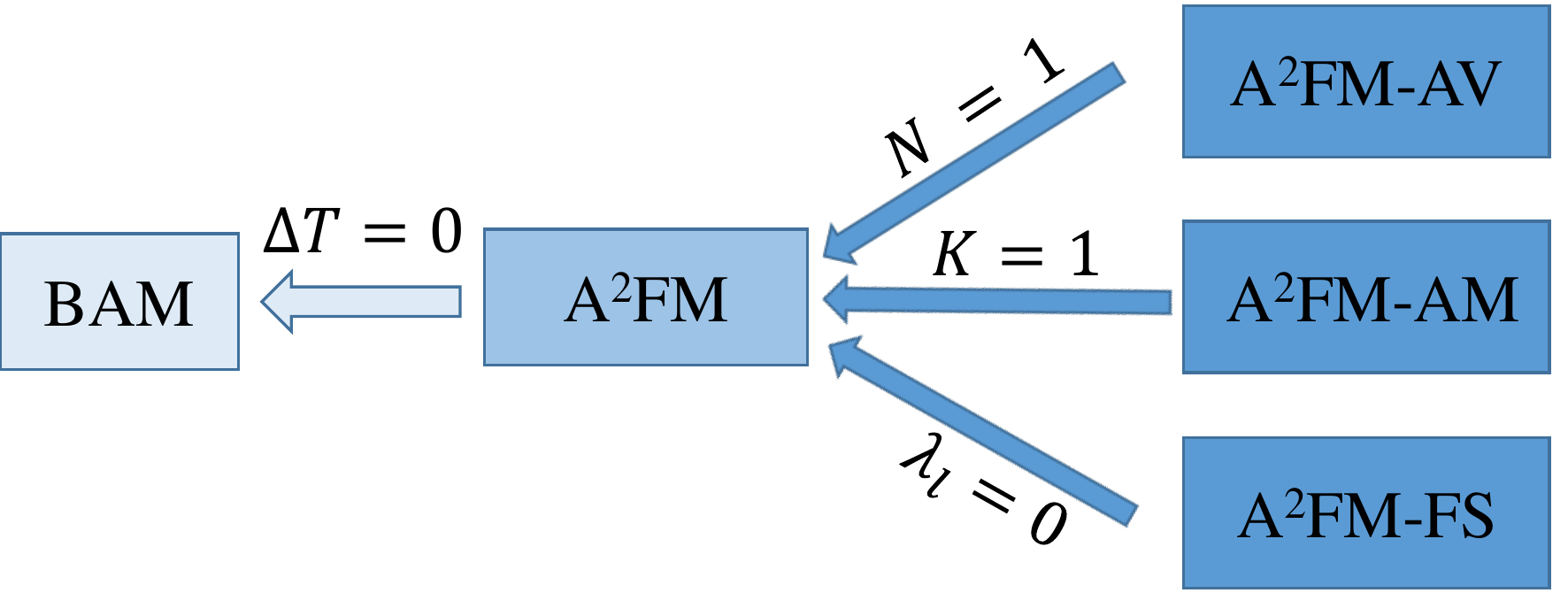}
\caption{Relationship between different attacks. Those attack models all related by the parameters like: the number of videos $N$, the number of models $K$, the number of adversarial frames $\Delta T$, and the weight of latent layer $\lambda_l$.}
\label{fig:figureMR}
\vspace{-2mm}
\end{figure}

We have introduced several attack models so far, and each model has its distinctive attack purpose.  At the end of this section, we show the relationships between different attack variants in Fig.~\ref{fig:figureMR}. The relationships can be summarized as follows:

\begin{itemize}
\vspace{-0.20cm}
\setlength{\itemsep}{-0.2ex} 
\item A\textsuperscript{2}FM-AV degrades to A\textsuperscript{2}FM if the number of videos $N$ = 1. This means A\textsuperscript{2}FM  focuses on a  single video while A\textsuperscript{2}FM-AV focuses on  multiple videos.

\item A\textsuperscript{2}FM-AM degrades to A\textsuperscript{2}FM if the number of models $K$ = 1. This means A\textsuperscript{2}FM attacks a single recognition model while ${A}^2$-AM-FM attacks multiple models.

\item A\textsuperscript{2}FM-FS degrades to A\textsuperscript{2}FM if the weight of latent layer $\lambda_l$ = 0. This means A\textsuperscript{2}FM-FS considers the feature representation similarity of intermediate layer while A\textsuperscript{2}FM does not.

\item  A\textsuperscript{2}FM degrades to basic attack method (BAM)(\textit{i.e.,} adding perturbations to the original video) if the number of adversarial frames $\Delta T$ = 0. This means A\textsuperscript{2}FM appends adversarial frames to the original video while BAM does not.
\end{itemize}

\section{Experiments}\label{sec:experiments}

In this section, we evaluate our proposed method comprehensively and compare our method with the state-of-the-art white-box video attack methods.

\subsection{Experimental Setting}\label{sec:setting}
\textbf{Datasets.} We use two popular benchmark datasets for video classification: UCF-101 \cite{soomro2012ucf101} and HMDB-51 \cite{kuehne2011hmdb}. UCF-101 is an action recognition dataset collected from YouTube, which contains 13320 realistic action videos within 101 action categories (\textit{e.g.} body-motion, sports, playing instruments, human-interaction ). Similarly, HMDB-51 is comprised of 6849 video clips distributed in 51 action categories including facial actions and body movements. Each category contains at least 101 video clips. 

\textbf{Video Recognition Models.} We consider up to six state-of-the-art video classification models, namely, I3D-Inception, I3D-ResNet, CNN+LSTM, C3D, ResNet3D and P3D, as our target models to attack. I3D is an inflated 3D convolutional network. The difference between I3D-ResNet and I3D-Inception lies at  the basic 2D model inflated to 3D. 
More specifically, the I3D-Inception model utilizes the UCF-101 dataset to fine-tune the pre-trained model~\cite{carreira2017quo},   
while I3D-ResNet is based on the  ResNet101 model trained on ImageNet. 
For CNN+LSTM, we use ResNet101 pretained on ImageNet as  feature extractor, and then train LSTM with features output from ResNet101. 
P3D and Resnet3D are all implemented officially. 
It should be noted that we only consider the RGB part for these video recognition models.
The accuracy of the six models on UCF-101 and HMDB-51 can be found in Tab.~\ref{tab:BAL}. The accuracy gap between ours and those reported in \cite{donahue2014longterm,Tran_2015,carreira2017quo,qiu2017learning,tran2018closer} is mainly caused by the availability of different input frames at test time.

\begin{table}[!tb]
\centering
  \caption{Basic accuracy ($\%$) of different video models.}
  \label{tab:BAL}
  \small
  \setlength{\tabcolsep}{1.8em} 
  \begin{tabular}{c c c }
    \toprule
     Models & UCF-101 & HMDB-51 \\
    \midrule
    I3D-ResNet & 57.7  & 97.2 \\
    I3D-Inception & 94.9 & 96.8 \\
    CNN+LSTM & 34.5 & 92.4 \\
    C3D & 50.9 & 99.9 \\
    ResNet3D & 83.7 & 91.6 \\
    P3D & 58.8 & 95.2 \\
    \bottomrule
  \end{tabular}
\vspace{-3mm}
\end{table}

\textbf{Attack Settings.} We insure every single video to be attacked is classified successfully. 
Each input video is set to have $T=24$ frames, and the length of the input video becomes 28 after attacking.
During the basic attack phase (attack a specific video under a specific model), we set the number of adversarial frames $\Delta T=2$, and each test video is expanded to 28 frames by appending another 2 frames from the original video. 
We select 500 videos from different categories in the test dataset to evaluate the attack performance for the different attack strategies in different attack scenarios. 
The parameters $\lambda$, $\alpha$ and $\beta$ are tuned in the training phase. 
All of the experiments are stopped when we find the adversarial perturbations or we reach the number of maximum iterations. 

To quantitative evaluate attack models, we  use the following performance measures. 
(i) Fooling Rate (FR): the ratio of the generated adversarial videos that are successfully misclassified. 
(ii) Average Absolute Perturbation (AAP): $AAP=\frac{1}{N \times S}\sum_{n=1}^N\frac{\sum|\mathbf{E}_n|}{\Delta T_n}$, where N is the number of test videos, 
and $S$ is the spatial size of perturbations ($S = 224 \times 224$).
$\mathbf{E}_n$ is the changed magnitude of perturbation for the $n$-th video, and  $\Delta T_n$ is the number of adversarial frames for the $n$-th video. Note that all of the experiments are based on the video classification models classify successfully. 
(iii) Difference between Intermediate Layer (DIFF): denotes the Euclidean distance  between the adversarial frames and the original frames at the  $l$-th intermediate layer.

\subsection{Performance and Transferability}
\begin{figure*}[!t]
\centering\includegraphics[width=0.92\textwidth]{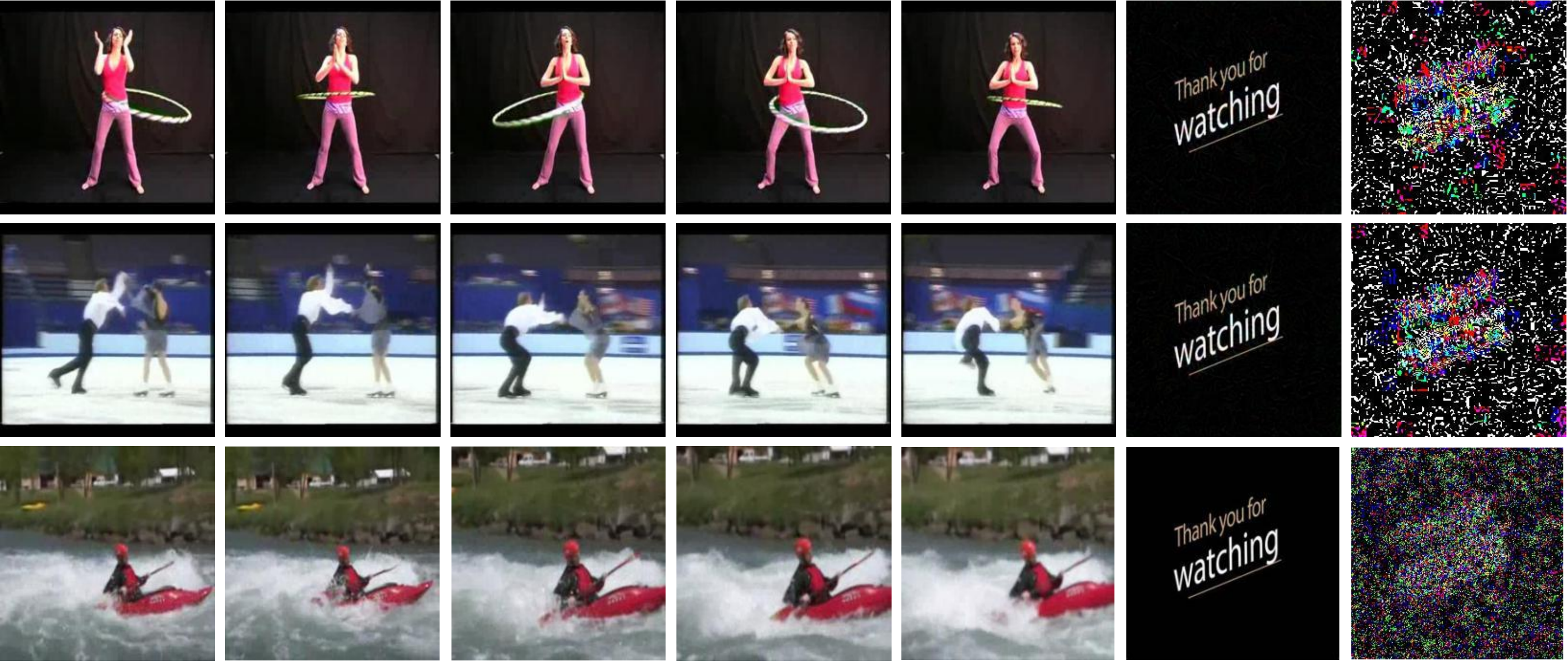}
\caption{Three adversarial videos are generated by A\textsuperscript{2}FM with Resnet3D. Top-5 columns are original videos and the sixth column is their corresponding adversarial frames. 
The last column is the perturbations. The last column shows the perturbations, in which the amplitudes are enlarged by 255 times for better visualization.}
\label{fig:figureBAE}
\vspace{-.2mm}
\end{figure*}

We first apply our attack method on a single network. Figure.~\ref{fig:figureBAE} shows two example adversarial videos generated by A\textsuperscript{2}FM (Eq.~\ref{eq:eq1}). 
The adversarial frame "Thank you for watching" is just like a normal ending frame, and adding perturbations does not make the adversarial frames strange. 
Table.~\ref{tab:basic} lists the performance of A\textsuperscript{2}FM on UCF-101 and HMDB-51. It indicates that $\mathbf{{A}^2}$FM outperforms BAM by a large gap on P, and meanwhile remains almost the same high FR as BAM. 
For example, with recognition model P3D, both A\textsuperscript{2}FM and BAM achieve perfect FR on UCF-101. 
However, the AAP of A\textsuperscript{2}FM is just 0.02, while the AAP of perturbation generated by \textbf{\cite{wei2018sparse}} is up to 0.2, which is 10 times compared with A\textsuperscript{2}FM. 
This means the proposed A\textsuperscript{2}FM  generates high quality adversarial videos  with more imperceptible perturbations.

\begin{table}[tb]
\centering
  \caption{Comparison of BAM and A\textsuperscript{2}FM with different video classification models.}
  \label{tab:basic}
  \small
  \setlength{\tabcolsep}{0.6em} 
  \begin{tabular}{c c rr rr}
    \toprule
  \multirow{2}{*}{Target Model} &\multirow{2}{*}{Methods} & \multicolumn{2}{c}{UCF-101} & \multicolumn{2}{c}{HMDB-51} \\
    \cmidrule{3-6}
    & & FR (\%)  & AAP & FR (\%)  & AAP \\
    \midrule
    \centering\multirow{2}{*}{I3D-ResNet}
    & BAM & \textbf{100} & 0.22 & \textbf{100} & 0.31 \\ 
    & A\textsuperscript{2}FM & \textbf{100} & \textbf{0.05} & \textbf{100} & \textbf{0.06} \\
    \hline
    \centering \multirow{2}{*}{I3D-Inception} & BAM & \textbf{99.5} & 0.20 & \textbf{100} & 0.28 \\
    & A\textsuperscript{2}FM & \textbf{99.5} & \textbf{0.08}  & \textbf{100} & \textbf{0.07} \\
    \hline
    \centering \multirow{2}{*}{CNN+LSTM} & BAM & \textbf{100} & 0.20 & \textbf{100} & 0.28\\
    & A\textsuperscript{2}FM & \textbf{100} & \textbf{0.02} & \textbf{100} & \textbf{0.02} \\
    \hline
    \centering \multirow{2}{*}{C3D} & BAM & \textbf{99.5} & 0.24 & \textbf{100} & 0.30 \\
    & A\textsuperscript{2}FM & 97.3 &  \textbf{0.14} & 96.8 & \textbf{0.16}\\
   \hline
    \centering \multirow{2}{*}{ResNet3D} & BAM & \textbf{97.8} & 0.25 & \textbf{100} & 0.30 \\
    & A\textsuperscript{2}FM & 95.1 & \textbf{0.09} & \textbf{100} & \textbf{0.07} \\
    \hline
    \centering \multirow{2}{*}{P3D} & BAM & \textbf{100} & 0.20 & \textbf{100} & 0.28 \\
    & A\textsuperscript{2}FM & \textbf{100 }& \textbf{0.02} & \textbf{100} & \textbf{0.02} \\
    \bottomrule
  \end{tabular}
\vspace{-2mm}
\end{table}

\textbf{Cross-Video Transferability.}
The experimental results in Tab.~\ref{tab:basic} show that the A\textsuperscript{2}FM can increase the effectiveness of adversarial perturbations for specific videos. However, we argue that a successful adversarial perturbation should not only perform well for a specific video, but also should hold the capability in transferring across different videos.
We evaluate the transferability of adversarial perturbations across videos on UCF-101 and HMDB-51 by  A\textsuperscript{2}FM-AV (Eq.~\ref{eq:eq3} ). 
The performance of universal adversarial perturbations  across videos is listed in Tab.~\ref{tab:CVT}.
One can clearly see that, compared with the baseline BAM, A\textsuperscript{2}FM-AV has much superior performance in FR. 
For example, with P3D as the treat model, the FRs of BAM and A\textsuperscript{2}FM-AV on  UCF-101 are $20.7\%$ and $98.4\%$, respectively.
The result suggests that our method can significantly enhance the attack ability in transferring across different videos. 
Our method has larger AAP than BAM in this scenario.
This is mainly because our attack orientation is more uniform and less random, which helps the perturbation accumulate gradually without breaking the decision boundary of threat models.
One can refer to Sec.~\ref{sec:ASP} for more discussions.
\begin{table}[tb]
\centering
  \caption{Comparison of BAM and A\textsuperscript{2}FM-AV  in  transferability  across  different videos.}
  \label{tab:CVT}
  \small
  \setlength{\tabcolsep}{0.5em} 
  \begin{tabular}{c c rr rr}
    \toprule
  \multirow{2}{*}{Target Model} &\multirow{2}{*}{Methods} & \multicolumn{2}{c}{UCF-101} & \multicolumn{2}{c}{HMDB-51} \\
    \cmidrule{3-6}
    & & FR (\%)  & AAP & FR (\%)  & AAP \\
    \midrule
    \centering\multirow{2}{*}{I3D-ResNet} & BAM & 95.4 & 0.62 & 93.0 & 0.70 \\
    & A\textsuperscript{2}FM-AV & \textbf{98.1} & \textbf{0.52} & \textbf{97.8} & \textbf{0.60} \\
    
    \hline
    \centering \multirow{2}{*}{I3D-Inception} & BAM & 2.6 & \textbf{0.34} & 2.0 & \textbf{0.25} \\
    & A\textsuperscript{2}FM-AV & \textbf{69.3} & 1.25  & \textbf{2.3} & 0.84 \\
    
    \hline
    \centering \multirow{2}{*}{CNN+LSTM} & BAM & 18.1 & \textbf{0.09} & \textbf{69.6} & \textbf{0.13}\\
    & A\textsuperscript{2}FM-AV & \textbf{47.1} & 0.16 & 45.7 & 0.21 \\
    
    \hline
    \centering \multirow{2}{*}{C3D} & BAM & 97.9 &\textbf{ 0.75} & \textbf{98.0} & \textbf{0.68} \\
    & A\textsuperscript{2}FM-AV & \textbf{98.1} &  1.21 & 96.9 & 1.75 \\
    
   \hline
    \centering \multirow{2}{*}{ResNet3D} & BAM & 45.2 & \textbf{0.65} & 58.6 & \textbf{0.49} \\
    & A\textsuperscript{2}FM-AV & \textbf{96.6} & 1.21 & \textbf{94.1} & 0.79 \\
    
    \hline
    \centering \multirow{2}{*}{P3D} & BAM & 20.7 & \textbf{0.11} & 46.9 & 0.16 \\
    & A\textsuperscript{2}FM-AV & \textbf{98.4} & 0.25 & \textbf{97.4} & \textbf{0.15} \\
    
    \bottomrule
  \end{tabular}
\vspace{-2mm}
\end{table}

\textbf{Cross-Model Transferability.}
We also evaluate the transferability of perturbations across models with A\textsuperscript{2}FM-AM (Eq.~\ref{eq:eq4}). 
To the end, we use the evaluated six models to explore the across models perturbations.
Specifically, we use the Leave-One-Out ensemble method that excludes one model to produce perturbations, 
and then attack each model with the generated perturbations. The corresponding results are shown in Tab.~\ref{tab:CMT}.
As we have seen before, the table shows that our method does actually have a better performance in fooling those models. 
For example, with perturbations generated by ensemble models without the P3D model,  the fooling rates of attacking P3D for BAM and  A\textsuperscript{2}FM-AM are $15.4\%$  and $59.0\%$, respectively.
This means the  gain of our method over the baseline is over $40\%$ for the unseen threat model. 
We think the improvement may be caused by the constrained and unified attacking orientation of our method, which makes the perturbations stay far away from the classification boundary.
\begin{table*}[tb]
\vspace{-.1cm}
\caption{Comparison of BAM and A\textsuperscript{2}FM-AM in transferability across models on UCF-101 dataset.  The first column  indicates we use the  Leave-One-Out ensemble method that excludes one model to produce perturbations. For instance,`$-$I3D-ResNet' means the corresponding ensemble model excludes I3D-ResNet. The numbers in the 3-8 columns are the fooling rates ($\%$) for each attacked model.}  
\centering
\small
\setlength{\tabcolsep}{1.2em}
\begin{tabular}{ c c c c c c c c }
\toprule
Models  & Method    & I3D-ResNet & ResNet3D & P3D & I3D-Inception & C3D & CNN+LSTM\\
\midrule
\multirow{2}{*}{$-$I3D-ResNet }  &BAM     &$ 0$    &$ 78.7$   &$ 84.6$    &$ 87.8$     &$ 70.8$   &$56.2$  \\
     &  A\textsuperscript{2}FM-AM  & $\textbf{39.5}$   &$68.1$   &$97.4$    &$42.9$        &$85.4$   &$81.6$  \\
\hline
\multirow{2}{*}{$-$ResNet3D} &  BAM &$ 100$    &$ 0$   &$ 84.6$    &$ 87.8$     &$ 70.8$   &$38.9$  \\
  &  A\textsuperscript{2}FM-AM  &$89.5$    &$\textbf{6.4}$   &$97.4$    &$52.2$        &$85.4$   &$71.4$   \\
\hline
\multirow{2}{*}{$-$P3D}       &  BAM     &$ 100$    &$ 80.9$   &$ 15.4$    &$  87.8$  &$ 72.9$   &$58.8$   \\
&  A\textsuperscript{2}FM-AM     &$86.8$    &$74.5$   &$\textbf{59.0}$   &$50.0$        &$85.4$   &$83.7$   \\
\hline
\multirow{2}{*}{$-$I3D-Inception}   &  BAM  &$100$    &$83.0$   &$97.4$    &$0$   &$73.0$   &$61.1$  \\
&  A\textsuperscript{2}FM-AM    &$86.8$    &$78.7$   &$100$    &$\textbf{2.0}$        &$85.4$   &$50.0$  \\
\hline
\multirow{2}{*}{$-$C3D}      &  BAM    &$100$    &$83.0$   &$100$    &$90.0$        &$0$   &$64.7$  \\
 &  A\textsuperscript{2}FM-AM    &$92.1$    &$80.9$   &$100$    &$60.0$        & $\textbf{20.8}$    &$79.6$ \\
\hline
\multirow{2}{*}{$-$CNN+LSTM}   &  BAM    &$100$    &$80.9$   &$97.4$    &$97.8$     &$72.9$    &$35.7$ \\
 &  A\textsuperscript{2}FM-AM    &$89.5$    &$74.5$   &$100$    &$55.6$        &$85.4$    &$\textbf{77.6}$ \\
\bottomrule
\end{tabular}\label{tab:CMT}
\vspace{-4mm}
\end{table*}

\begin{figure}[!tb]
\centering\includegraphics[width=0.45\textwidth]{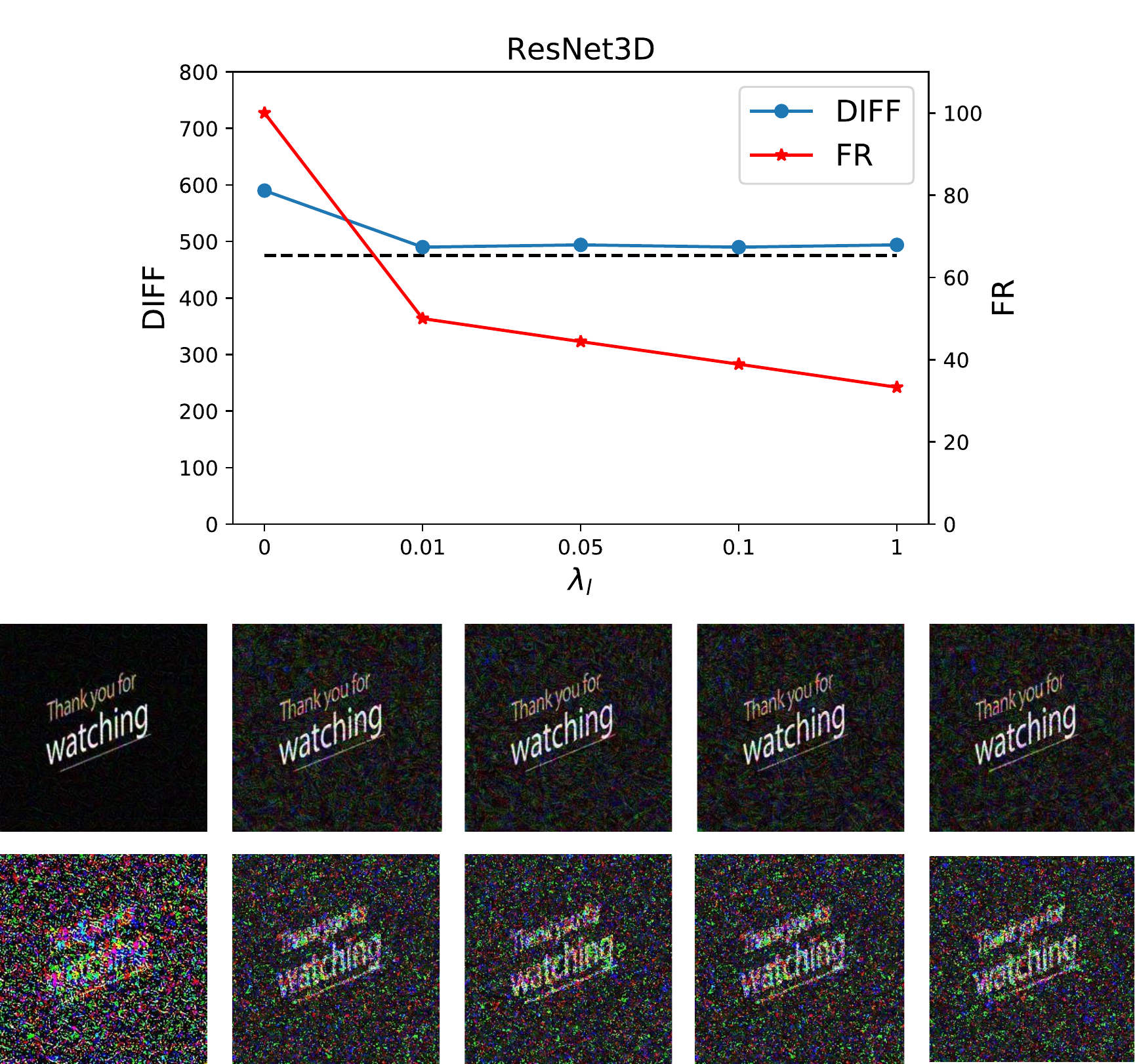}
\caption{An example of A\textsuperscript{2}FM-FS. The $x$-axis is the parameter $\lambda_l$, namely, the penalty on feature similarity in Eq.~\ref{eq:eq5}. The left $y$-label is the (\textbf{DIFF}) between clean video frames and adversarial video frames, and the right $y$-axis denotes the (\textbf{FR}). The black dotted line is the average normal distance between original frames. In the bottom of the figure, we show the corresponding adversarial frames that are appended to the original video. The first row is the adversarial frames with five $\lambda_l$ values (From left to right: $\lambda_l=0, 0.01, 0.05, 0.1, 1$), and the second row is the corresponding adversarial perturbation.}
\label{fig:figureAFR}
\vspace{-.4cm}
\end{figure}

\textbf{Attack with Feature Similarity.}\label{sec:AFR}
Despite the state-of-the-art performance of Eqs.~\ref{eq:eq1},~\ref{eq:eq3} and~\ref{eq:eq4}, 
there are relatively large distances between the adversarial frames and the original frames at intermediate layers, which may cause defenders find adversarial frames easily. 
Our method  A\textsuperscript{2}FM-FS (\textit{i.e.}, Eq.~\ref{eq:eq5}) can decrease the perceptibility of adversaries effectively.
As an example, we validate A\textsuperscript{2}FM-FS with ResNet3D on UCF-101, where we choose the last convolutional layer as feature representations of frames. 
The results are displayed in Fig.~\ref{fig:figureAFR}. As shown, DIFF is falling fast with the regularization parameter $\lambda_l$ increased from 0 to 0.01, then it keeps stable and nearly approaches the benchmark (refers to the average normal distance between the original frames) with  $\lambda_l$ from 0.01 to 1. This means when $\lambda_l>0.01$ the adversarial frames are similar to the original frames at the internal layer of the deep network, and therefore it is hard for a simple defender to perceive adversarial frames. 
As expected,  FR decreases gradually with the increase of penalty $\lambda_l$ on feature similarity.   
Our method suggests that one can choose a suitable $\lambda_l$ (For instance, we can set $\lambda_l=0.01$ for the given example) to balance the attacker and the defender.

\begin{figure}[!tb]
\centering\includegraphics[width=0.45\textwidth]{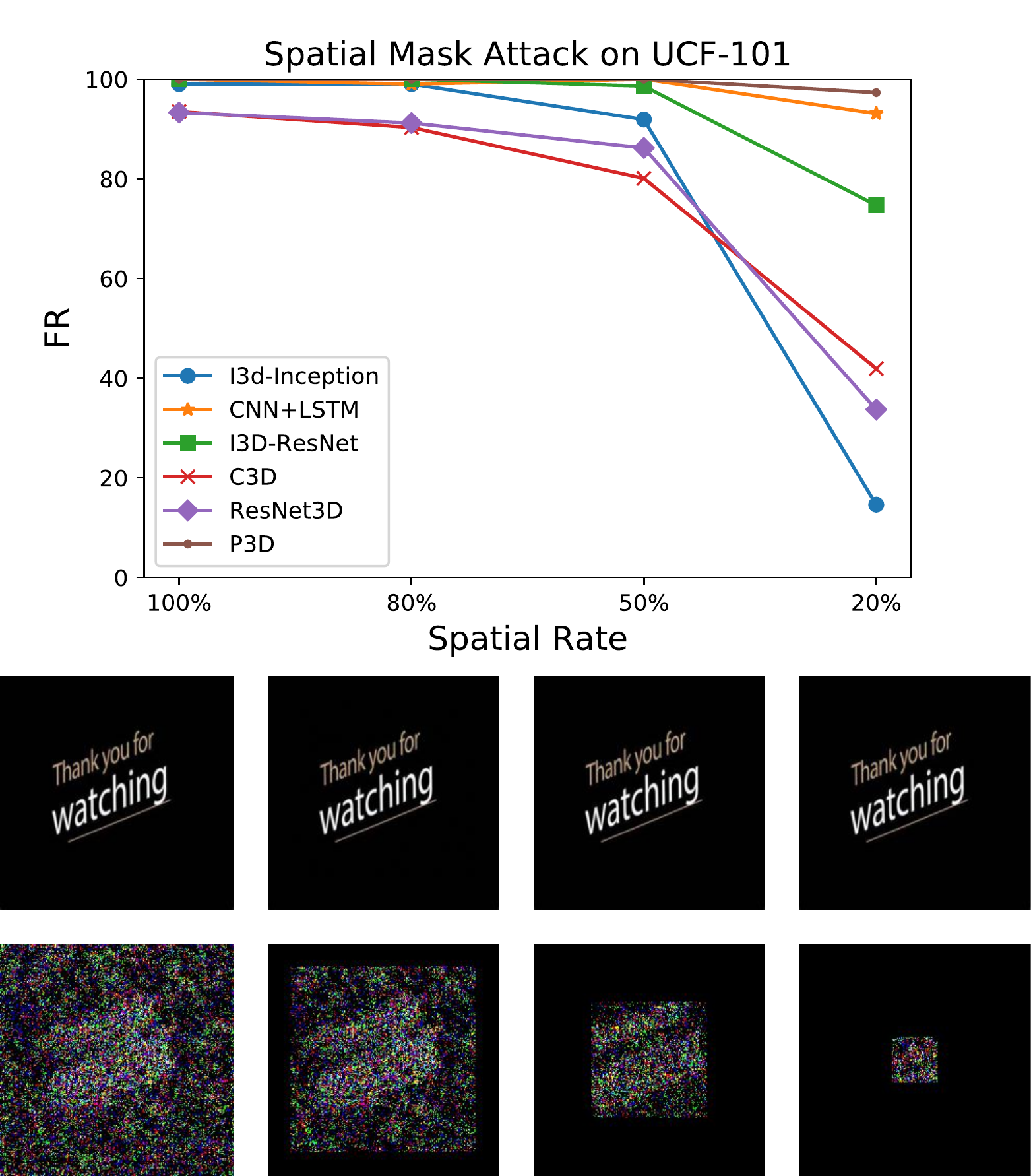}
\caption{Illustration the (\textbf{FR}) values with different spatial rates. We here report the results when spatial rates are $100\%$, $80\%$, $50\%$, $20\%$, respectively. To be simple, the spatial masks are constrained to be squares. In the bottom of the figure, the first row is example adversarial frames of attacking ResNet3D and the second row is the corresponding perturbations. } 
\label{fig:figureSMA}
\vspace{-4mm}
\end{figure}
 
\subsection{Other Attack Options}
\textbf{Spatial Mask Attack.}\label{sec:SMA}
With aforementioned methods, each pixel of the adversarial frames is perturbed as the spatial size of the perturbations is the same as the adversarial frames. This may increase the risk to perceive adversarial frames. To increase the concealing chance, one can restrict the perturbation in a much smaller size. 
By using spatial masks to filter out the perturbations in some positions, we could generate more imperceptible adversarial frames.
It is quite easy for us to construct arbitrary shape perturbations by changing shapes of spatial masks. However, to be simple, we only show the fooling rates for square perturbations in Fig.~\ref{fig:figureSMA}.
As the figure shown, the FR values decrease slowly when the spatial rate drops from $100\%$ to $50\%$. This means spatial mask attack is indeed effective for generating more imperceptible adversarial frames while slightly or not hurting FR with a suitable spatial rate. 
Furthermore, it is easy to understand that when the spatial rate is too small, the generated perturbations can not have enough ability to attack classification models. For instance, as shown, the FR drops a lot when the spatial rate equals to $20\%$ 


\textbf{Targeted Attack.}\label{sec:TA}
As mentioned in Eq.~\ref{eq:eq2}, we evaluate the performance of our method attack specified target labels. We separately choose 5 different labels as our target attack labels from UCF-101 and HMDB-51. With each target label, we set other categories as testing set. Then, we randomly choose one video for each test class, and thus, we obtain a new testing set with 100 videos in UCF-101 and another testing set with 50 videos in HMDB-51.
The experimental results are shown in Tab.~\ref{tab:TAT}.
As we can see, our method is inferior to BAM in terms of FR for some cases.  This is mainly because the targeted attack needs a specific gradient direction to proceed, so the constrained attack direction may not help the adversarial attack accurately to find the right way to achieve a specific classification region. 
However, with the same or nearly the same fooling rate, A\textsuperscript{2}FM usually has smaller AAP than BAM, which further demonstrates A\textsuperscript{2}FM can actually help to generate smaller adversarial perturbation in basic attack settings. 
\begin{table}[tb]
\centering
  \caption{Comparison of BAM and A\textsuperscript{2}FM  for targeted attack.}
  \label{tab:TAT}
  \small
  \setlength{\tabcolsep}{0.6em} 
  \begin{tabular}{c c rr rr}
    \toprule
  \multirow{2}{*}{Target Model} &\multirow{2}{*}{Methods} & \multicolumn{2}{c}{UCF-101} & \multicolumn{2}{c}{HMDB-51} \\
    \cmidrule{3-6}
    & & FR (\%)  & AAP & FR (\%)  & AAP \\
    \midrule
    \centering\multirow{2}{*}{I3D-ResNet} & BAM & 97.6 &  0.29 & \textbf{97.8} &  0.31 \\
    & A\textsuperscript{2}FM & \textbf{97.7} & \textbf{0.17} & \textbf{97.8} & \textbf{0.14} \\
    \hline
    \centering \multirow{2}{*}{I3D-Inception} & BAM & \textbf{84.6} &  0.23 & \textbf{96.8} &  0.27 \\
    & A\textsuperscript{2}FM & 27.4 & \textbf{0.08} & 40.2 & \textbf{0.08} \\
    \hline
    \centering \multirow{2}{*}{CNN+LSTM} & BAM & \textbf{61.6} &  0.23 & \textbf{55.8} &  0.27 \\
    & A\textsuperscript{2}FM & 53.2 &\textbf{ 0.07} & 42.4 & \textbf{0.07}\\
    \hline
    \centering \multirow{2}{*}{C3D} & BAM & \textbf{97.9} &   0.30 & \textbf{97.8} &  0.31\\
    & A\textsuperscript{2}FM & 83.8 & \textbf{0.26} & 95.0 & \textbf{0.22} \\
   \hline
    \centering \multirow{2}{*}{Resnet3D} & BAM & \textbf{98.1} &  0.28 & \textbf{98.0} &  0.30 \\
    & A\textsuperscript{2}FM & \textbf{98.1} & \textbf{0.15} & \textbf{98.0} & \textbf{0.13} \\
    \hline
    \centering \multirow{2}{*}{P3D} & BAM & \textbf{98.0} &  0.22 & \textbf{97.8} &  0.26 \\
    & A\textsuperscript{2}FM & 97.8 & \textbf{0.07} & \textbf{97.8} & \textbf{0.08} \\
    \bottomrule
  \end{tabular}
\vspace{-4mm}
\end{table}

\section{Why is Our Attack Universal?}\label{sec:ASP}

\begin{figure}[!tb]
\centering\includegraphics[width=0.45\textwidth]{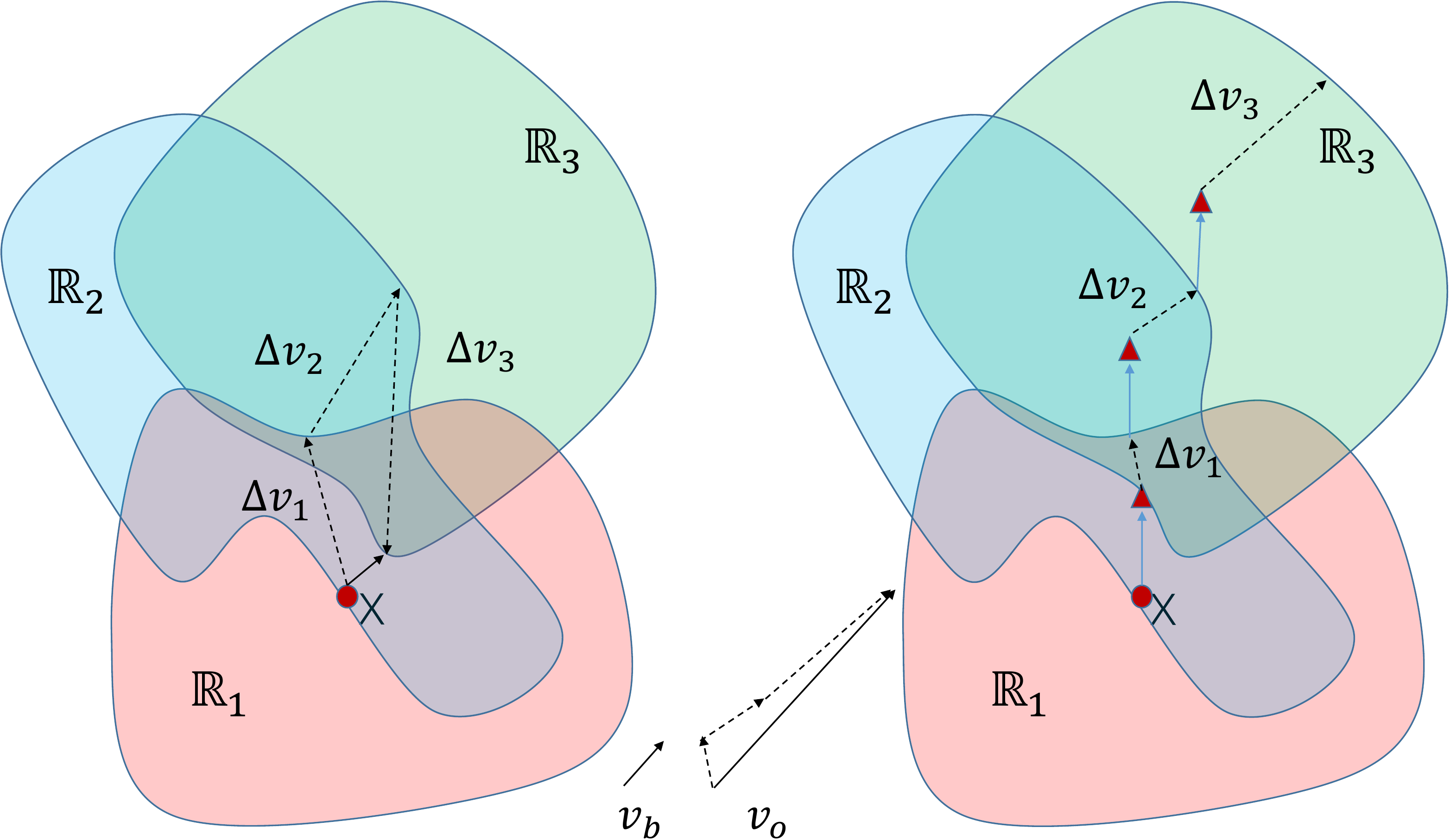}
\caption{An explanation of the mechanism of existing methods and our attacking method. The left figure shows the schematic of existing methods, while the right figure shows the schema represented by our method, where $v_b$ and $v_o$ denote the universal perturbation generated by the two methods, respectively. $X$ with red dot denotes the batch of video data and the red triangle denotes the video data after appending the adversarial frames. As for the classification region $\mathbb{R}_i$, they are showed in different shape with different colors. The blue arrow line in the classification region denotes the shift that are caused by the adversarial frames appended to the video, and the black dotted arrow line represents the minimal perturbation $\Delta v_i$ at each iteration. Finally, the black arrow line denotes the final universal perturbation we computed. }
\label{fig:figureASP}
\vspace{-4mm}
\end{figure}

The purpose of this section is to explain the strength of our method compared with BAM in universal perturbations. 
We empirically assume that, by constraining the diversity of perturbations, our method can generate a more unify perturbations  for adversarial frames. 
As illustrated in Fig.~\ref{fig:figureASP}, we show how perturbations are generated gradually  for both BAM (left figure) and our method (right figure) when the iteration goes on. At each iteration, the minimal perturbation $\Delta v_i$ is solved to reach the classification boundary. 

We can see the difference between two methods. 
BAM can not constrain the attack orientation as it chooses uncertain gradient ascent direction. With the stochastic attack orientation, there comes out the possibility of canceling each attack orientation out,  which of course makes the perturbations small. 
As the figure shown, small perturbations likely can not fight against or stay far away decision boundary, and therefore consequently result in a weak universal perturbation. 
However, with the setting of appending adversarial frames, we can make the attack orientation more uniform and similar, which helps a lot in choosing a more reasonable attack orientation.
As the figure shown, our method constrains the attack orientation certainly and reduces the chance of counteracting the attack contribution, therefore our method is more likely to find  robust universal perturbations and consequently has strong transfer ability.

\section{Conclusions}

In this paper, we present an interesting idea of attacking videos. We take advantage of the temporal property of videos, \textit{i.e.}, changing a few ending frames of it, though introducing a large perturbation in terms of the pixel-wise metrics, can still be easily concealed, \textit{i.e.}, most people would not notice that the video has been attacked. On the other hand, by adding adversarial perturbations \textit{only} on these new frames, the perceptibility of the added noise becomes smaller yet the attack is verified easier to transfer across videos and even different networks. In other words, our method, though simple, provides an effective pipeline of universal video attack.

This work inspires future research in the aspect of adding adversaries to 3D/4D image data. In particular, it remains unclear if the attacking methods on videos (2D spatial \& 1D temporal) are generalized to medical images (3D spatial) or medical videos (3D spatial \& 1D temporal). Also, it raises the issue of defending such new kind of attacks, which can be of importance for security reasons.

{\small
\bibliographystyle{ieee_fullname}
\bibliography{egbib}
}

\clearpage
\appendix
\section{Supplementary Material}
This Supplementary Material provides additional details and experimental results which we did not mentioned before.
\begin{itemize}
\vspace{-0.20cm}
\setlength{\itemsep}{-0.2ex} 
\item In Sec.~\ref{sec:AD}, we provide model details in attacking phase (\textit{e.g. the parameters of attack setting}).
\item In Sec.~\ref{sec:TD}, we provide more training details for reproducing video classification models. 
\item In Sec.~\ref{sec:DAF}, we provide the results of A\textsuperscript{2}FM under different adversarial frames to prove that the attack performance of our method does not influenced by the pattern of adversarial frames, which are shown in Tab.~\ref{tab:A2FM}. 
\item In Sec.~\ref{sec:SSMA}, we provide additional results on UCF-101 dataset under a special spatial mask attack setting to demonstrate that we can generate more imperceptible adversarial examples, which are shown in Tab.~\ref{tab:FSM} and the corresponding visualization are shown in Fig.~\ref{fig:3SM}.
\item In Sec.~\ref{sec:ACMH5}, we provide the results of the transferability of perturbations across models with A\textsuperscript{2}FM-AM on HMDB-51, which are shown in Tab.~\ref{tab:ATMT}. 
\item In Sec.~\ref{sec:DIFF}, we provide additional results of feature similar attack setting in different models, which are shown in Fig.~\ref{figure:DIFF}.
\end{itemize}

\subsection{Attacking Details}\label{sec:AD}
The max number of iterations for BAM and A\textsuperscript{2}FM is 40, 
while the max number of iterations for  A\textsuperscript{2}FM-AV and A\textsuperscript{2}FM-AM is $10 \times N$ (the number of testing videos) and $5 \times K$ (the number of ensemble models), respectively. 
For these four models, the threshold $\epsilon$ for magnitude of adversarial perturbations is $0.001$. 
For A\textsuperscript{2}FM, the feature extractor to measure the distance of internal layer representation is ResNet50. The max number of iterations is 200 and the step size of perturbation $\epsilon = 0.01$.

\subsection{Training Details}\label{sec:TD}
The evaluated models are trained on a workstation with 4 GeForce GPUs (each 11GB memory).
At the data pre-processing stage, the input frames of videos are resized to $224\times224$ and the value range is transformed from [0,255] to [0,1]. We only use the first 28 frames of videos for model training and evaluation.
We randomly divide the trimmed videos into a training set and test set, where the ratio of the number of training examples and the number of test examples is $9:1$.
We use one-hot encoding to represent different classes.  There are some differences among models: 
\begin{itemize}
\vspace{-0.20cm}
\setlength{\itemsep}{-0.2ex} 
\item C3D: This model used in our paper contains 3D convolutional layers, and followed by batch normalization layer with RELU activation. We set 0.2 as dropout rate to avoid overfitting and set the learning rate to 1e-4. The batch size is 16 and we trained it 15 epochs totally.
\item CNN+LSTM: This model contains two parts. The first part is a normal 2D convolutional network (ResNet50) in our paper. Then we use the LSTM model to copy with the temporal domain. Due to the restriction of memory, we set batch size 5 here and use 40 epochs totally to train it.
\item I3D-ResNet: The base model is ResNet50 and the batch size is 16. We train it 15 epochs totally.
\item I3D-Inception: This model is the same as I3D-ResNet. The difference is that the base model is Inception model. Due to low rate of convergence, we load the pre-trained model and fine-tuned it at the target dataset. 
\item ResNet3D: This model is similar to C3D, and its base model is ResNet50. We load the 2D parameters pre-trained on ImageNet for the ResNet50.
\item P3D: We use P3D63 in our experiment. The batch size is 10 here and we train it for 15 epochs.
\end{itemize}

\begin{table}[!t]
\centering
  \caption{Performance of A\textsuperscript{2}FM under different adversarial frames on UCF-101.}
  \label{tab:A2FM}
  \small
  \setlength{\tabcolsep}{0.4em} 
  \begin{tabular}{c rr rr rr}
    \toprule
  \centering Target Model  & \multicolumn{2}{c}{TFW1} & \multicolumn{2}{c}{TFW2} & \multicolumn{2}{c}{TFW3}\\
    \cmidrule{2-7}
    & FR (\%)  & AAP & FR (\%)  & AAP & FR (\%)  & AAP \\
    \midrule
    \centering I3D-ResNet
     & 100 & 0.04 & 100 & 0.04 & 100 & 0.05\\
    \hline
    \centering I3D-Inception 
     & 100 & 0.09  & 99.5 & 0.09  & 99.5 & 0.10\\
    \hline
    \centering CNN+LSTM 
    & 100 & 0.02 & 98.9 & 0.02 & 100 & 0.02\\
    \hline
    \centering C3D 
     & 95.2 &  0.15 & 96.3 & 0.15 & 97.3 & 0.17\\
   \hline
    \centering ResNet3D 
     & 97.4 & 0.09 & 96.8 & 0.09  & 97.3 & 0.10\\
    \hline
    \centering P3D 
     & 100 & 0.02 & 100 & 0.02 & 100 & 0.02\\
    \bottomrule
  \end{tabular}
\end{table}
\begin{table}[!t]
\centering
  \caption{Performance of A\textsuperscript{2}FM with special perturbation under Different Adversarial Frames on UCF-101. The percentage in brackets is the spatial rate of the spatial mask. For instance, `TFW1 ($16\%$)' denotes that the number of changed pixels occupies $16\%$ in the whole pixels of the adversarial frame `TFW1'. }
  \label{tab:FSM}
  \small
  \setlength{\tabcolsep}{0.3em} 
  \begin{tabular}{c rr rr rr}
    \toprule
  \centering Target Model  & \multicolumn{2}{c}{TFW1 ($16\%$)} & \multicolumn{2}{c}{TFW2 ($16\%$)} & \multicolumn{2}{c}{TFW3 ($11\%$)}\\
    \cmidrule{2-7}
    & FR (\%)  & AAP & FR (\%)  & AAP & FR (\%)  & AAP \\
    \midrule
    \centering I3D-ResNet
     & 74.7 & 0.001 & 82.4 & 0.001 & 84.8 & 0.001\\
    \hline
    \centering I3D-Inception 
     & 1.5 & 0.003  & 4.0 & 0.003  & 2.0 & 0.002\\
    \hline
    \centering CNN+LSTM 
    & 88.6 & 0.001 & 87.8 & 0.001 & 84.8 & 0.001\\
    \hline
    \centering C3D 
     & 28.8 &  0.003 & 6.8 & 0.004 & 25.5 & 0.002\\
   \hline
    \centering ResNet3D 
     & 14.3 & 0.004 & 17.6 & 0.003  & 19.4 & 0.003\\
    \hline
    \centering P3D 
     & 91.4 & 0.003 & 86.2 & 0.001 & 86.5 & 0.001\\
    \bottomrule
  \end{tabular}
\end{table}
\begin{table*}[!t]
\caption{Comparison of BAM and A\textsuperscript{2}FM-AM in transferability across models on HMDB-51 dataset.  The first column  indicates we use the  Leave-One-Out ensemble method that excludes one model to produce perturbations. For instance,`$-$I3D-ResNet' means the corresponding ensemble model excludes I3D-ResNet. The numbers in the 3-8 columns are the fooling rates ($\%$) for each attacked model.} 
\centering
\setlength{\tabcolsep}{0.8em} 
\begin{tabular}{c c c c c c c c }
\toprule
Models  & Method    & I3D-ResNet & ResNet3D & P3D & I3D-Inception & C3D & CNN+LSTM\\
\midrule
$-$I3D-ResNet   &BAM  &$2.1 $    &$ 91.5 $   &$ 100 $    &$ 6.1 $     &$ 64.0 $   &$63.0 $  \\
     &  A\textsuperscript{2}FM-AM  & $\mathbf{19.1 }$   &$95.7 $   &$100 $    &$2.0 $        &$58.0 $   &$61.7 $  \\
\hline
$-$ResNet3D &  BAM &$ 100 $    &$2.1 $   &$ 100 $    &$ 12.2 $     &$ 64.0 $   &$67.4 $  \\
  &  A\textsuperscript{2}FM-AM  &$100 $    &$\mathbf{6.4 }$   &$100 $    &$6.1 $        &$58.0 $   &$61.7 $   \\
\hline
$-$P3D       &  BAM     &$ 100 $    &$ 87.2 $   &$ 15.2 $    &$ 6.1 $  &$ 64.0 $   &$60.4 $   \\
&  A\textsuperscript{2}FM-AM     &$100 $    &$93.6 $   &$\mathbf{93.5 }$   &$2.0 $        &$58.0 $   &$60.0 $   \\
\hline
$-$I3D-Inception   &  BAM  &$100 $    &$87.2 $   &$91.3 $    &$\mathbf{0.0} $   &$64.0 $   &$54.2 $  \\
&  A\textsuperscript{2}FM-AM    &$100 $    &$93.6 $   &$100 $    &$\mathbf{0.0} $        &$48.0 $   &$60.4 $  \\
\hline
$-$C3D      &  BAM    &$100 $    &$87.2 $   &$87.0 $    &$90.0 $        &$0.0 $   &$54.2 $  \\
 &  A\textsuperscript{2}FM-AM    &$100 $    &$93.6 $   &$100 $    &$4.1 $        & $\mathbf{4.0 }$    &$62.5 $ \\
\hline
$-$CNN+LSTM   &  BAM    &$100 $    &$87.2 $   &$100 $    &$6.1 $     &$64.0 $    &$36.7 $ \\
 &  A\textsuperscript{2}FM-AM   &$100 $    &$93.6 $   &$100 $    &$2.0 $        &$58.0 $    &$\mathbf{44.7 }$ \\
\bottomrule
\end{tabular}
\label{tab:ATMT}
\end{table*}


\subsection{A\textsuperscript{2}FM under Different Adversarial Frames}\label{sec:DAF}
The goal of this section is to investigate the relationship between the adversarial frames and attack performance of A\textsuperscript{2}FM. Those different adversarial frames are shown in the first row of Fig.~\ref{fig:3SM}, which defined as 'TFW1`, `TFW2', and `TFW3', respectively. They represent different patterns. The left figure is a zoom in version of the adversarial frame we mentioned before, the middle is the adversarial frame with a white background which is the reverse of the former adversarial frame and the right one is an adversarial frame with different font size and different font style. These new appending frames are shown in very different patterns and the attack performance from these frames is so extreme that can cover all possible of existing adversarial frames. 
See Tab.~\ref{tab:A2FM} for the performance of A\textsuperscript{2}FM under different adversarial frames on UCF-101. There is nearly no performance gap between different adversarial frames, and therefore it is easy to conclude that the patterns of ending frames will not influence the performance of A\textsuperscript{2}FM.

\subsection{Special Spatial Mask Attack}\label{sec:SSMA}
\begin{figure*}[htbp]
\centering\includegraphics[width=0.8\textwidth]{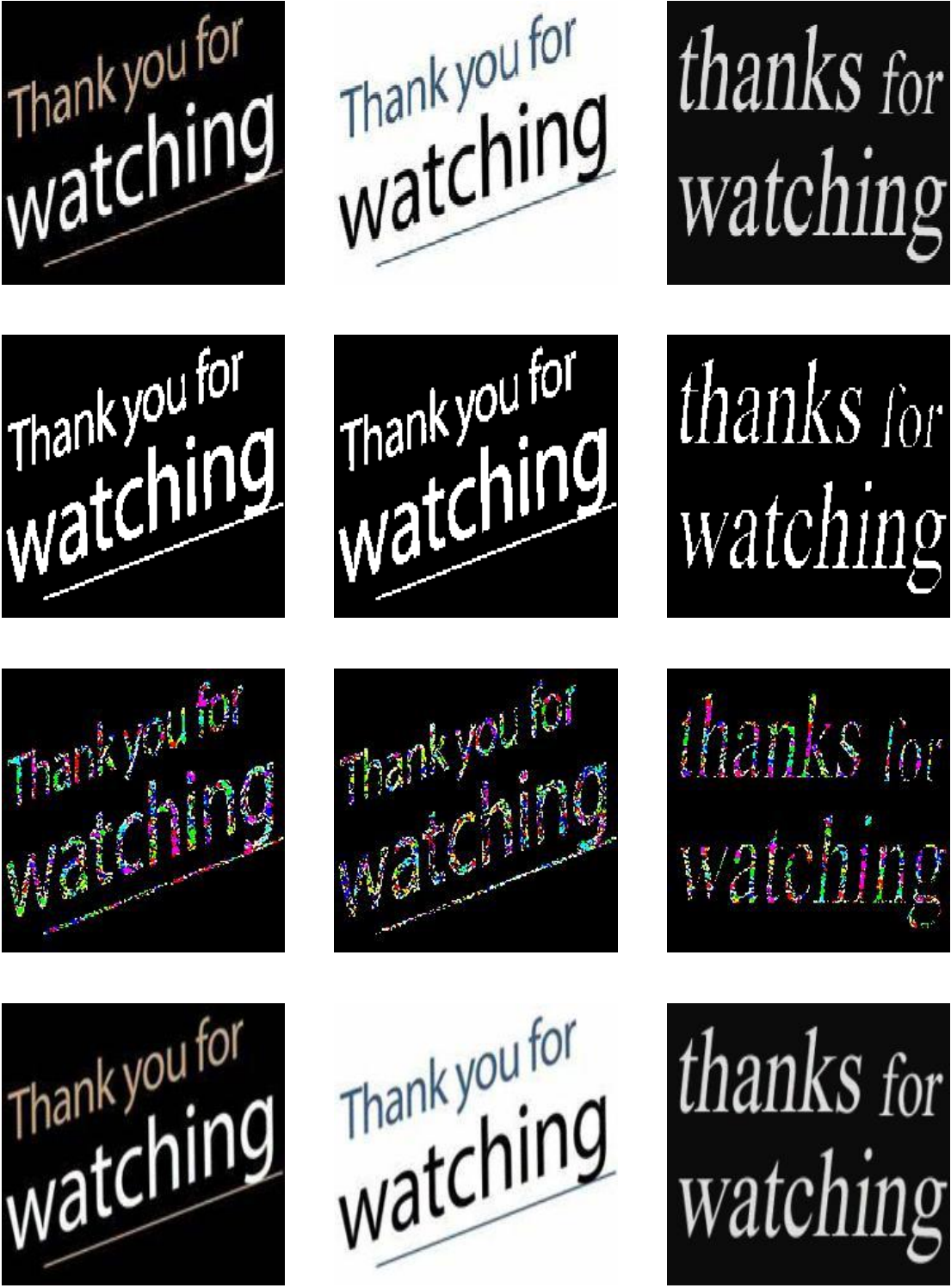}
\caption{Three examples of adversarial frames and its corresponding spatial mask, adversarial frames and spatial perturbations. The first row is adversarial frames without perturbation. The second row is the corresponding spatial mask for filtering the background to make the attack focus on the font. The third row represents the corresponding perturbation with a certain font spatial mask (amplify with $\times 255$ for better visualize). The last row is the final adversarial frames which we appending to original videos.}
\label{fig:3SM}
\vspace{-4mm}
\end{figure*}

As we mentioned before, we can construct arbitrary shape perturbations by changing shapes of spatial masks to generate a more imperceptible perturbation. Empirically speaking, the perturbation adding to the abundant texture areas can make the perturbation more imperceptible, so we make the mask that filters the background and keeps the font to be attacked. We visualize three examples of adversarial frames and their corresponding spatial masks, adversarial frames as well as spatial perturbations (as shown in Fig.~\ref{fig:3SM}). See Tab.~\ref{tab:FSM} for more detail in attacking a fixed model with a specific spatial perturbation.  

\subsection{Experimental Results on HMDB-51}\label{sec:ACMH5}
We provide the results of the transferability of perturbations across models with A\textsuperscript{2}FM-AM on UCF-101 as mentioned in Sec.~4.2. And the results shown in Tab.~\ref{tab:ATMT} indicate the attack transferability across models on HMDB-51 dataset remains robust and powerful.

\subsection{Performance in Feature Similar Attacking}\label{sec:DIFF}
\begin{figure*}[tb]
    \centering
        \subfigure{\includegraphics[width=0.45\hsize]{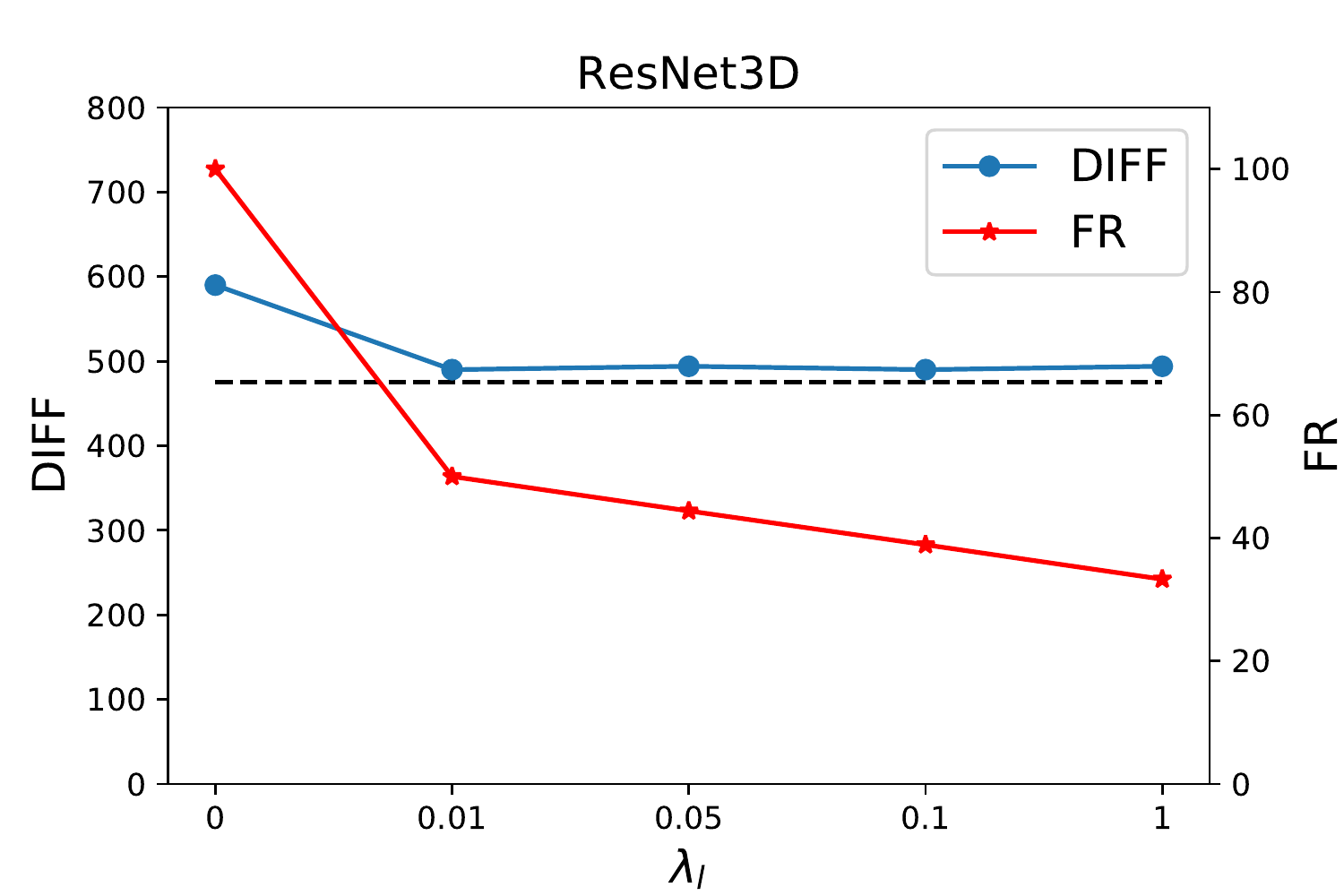}}  
        \subfigure{\includegraphics[width=0.45\hsize]{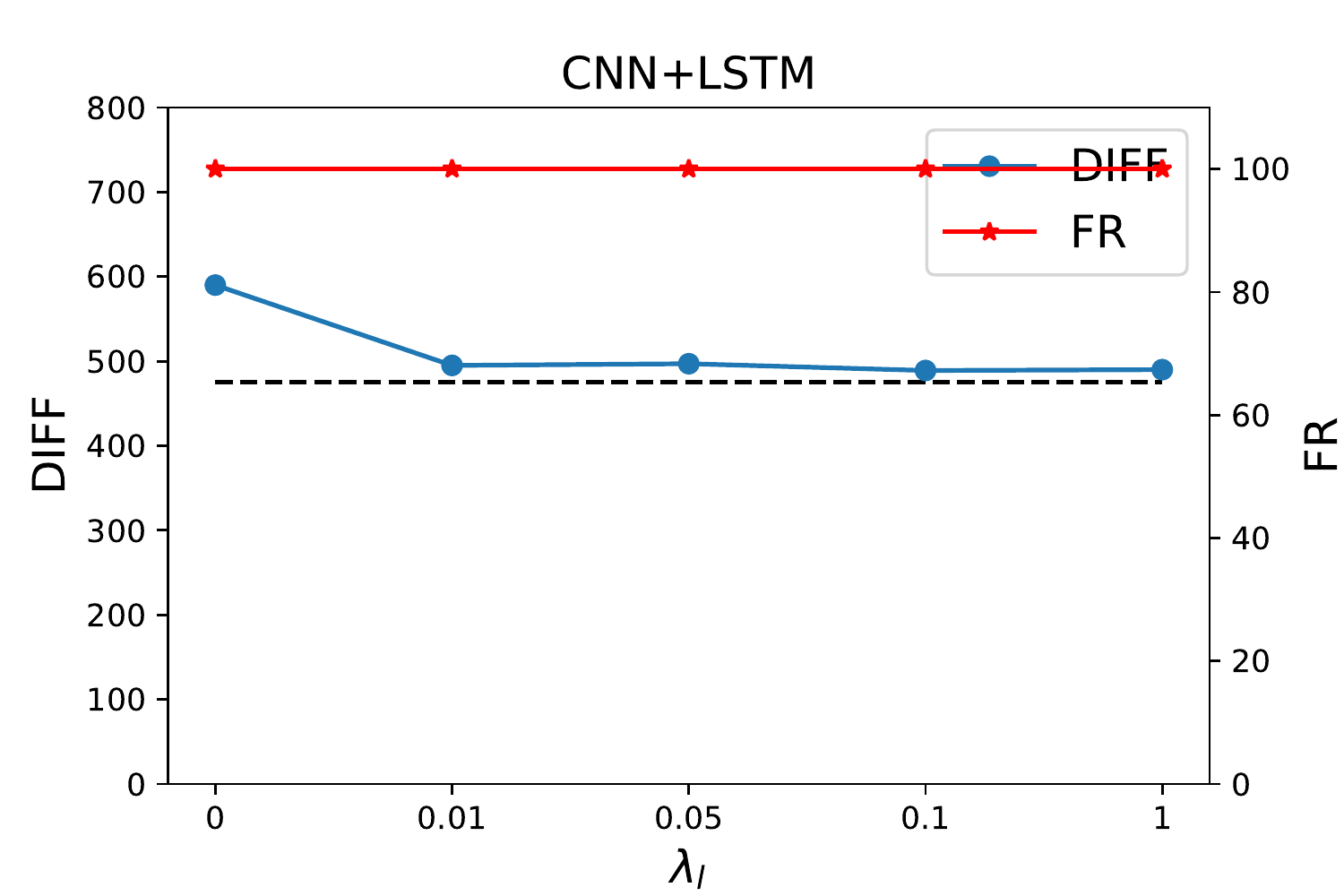}} \\
        \subfigure{\includegraphics[width=0.45\hsize]{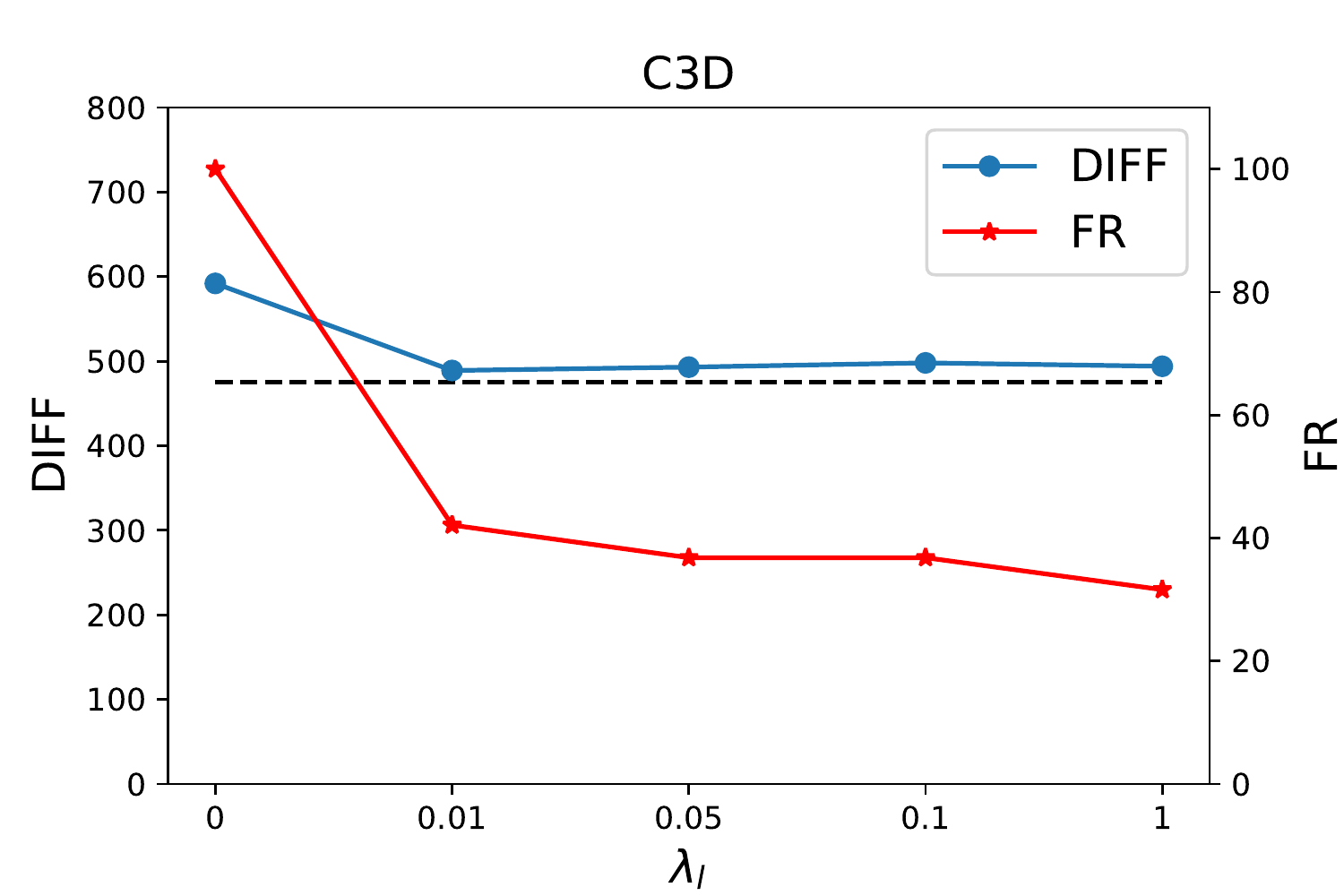}} 
        \subfigure{\includegraphics[width=0.45\hsize]{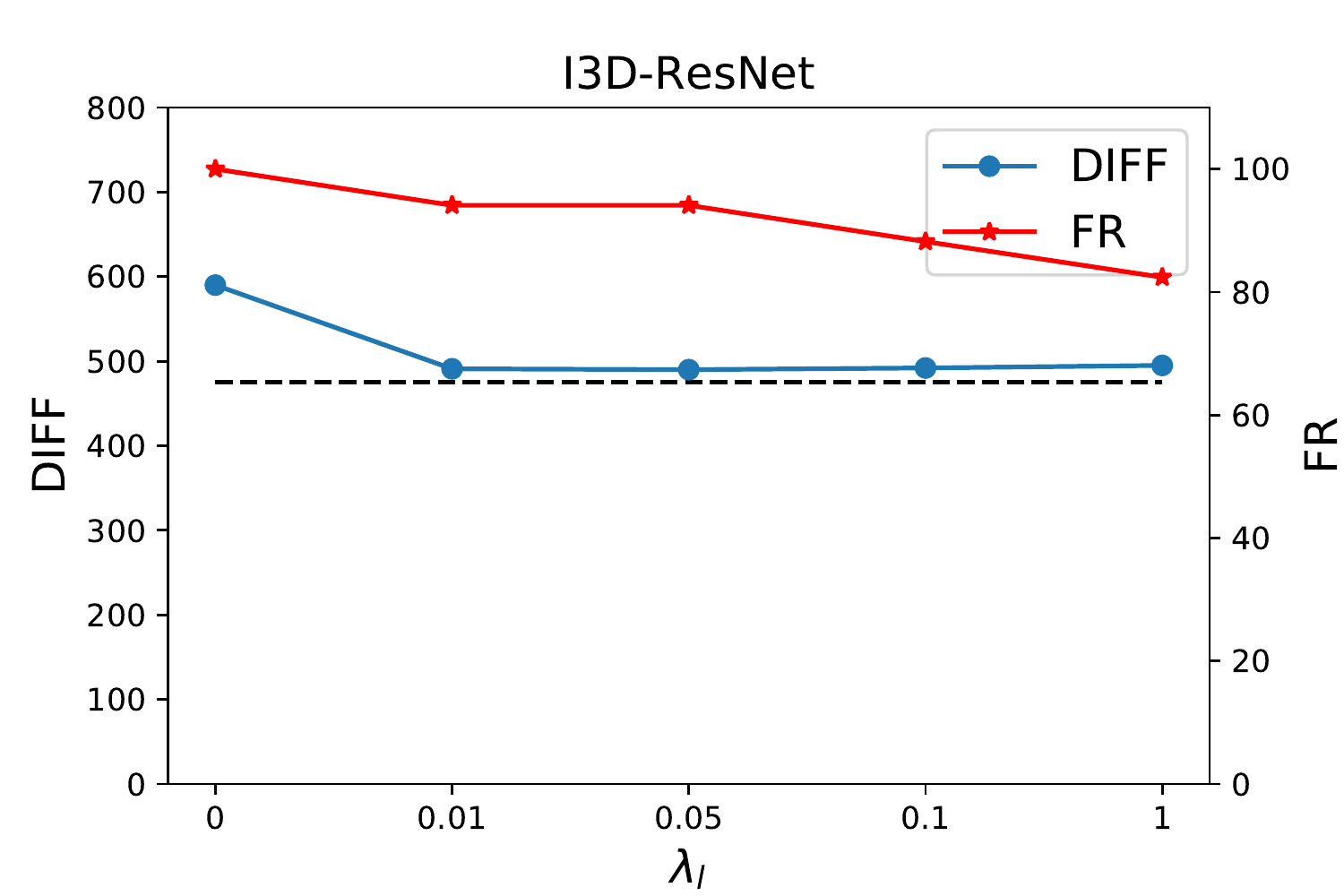}} \\
        \subfigure{\includegraphics[width=0.45\hsize]{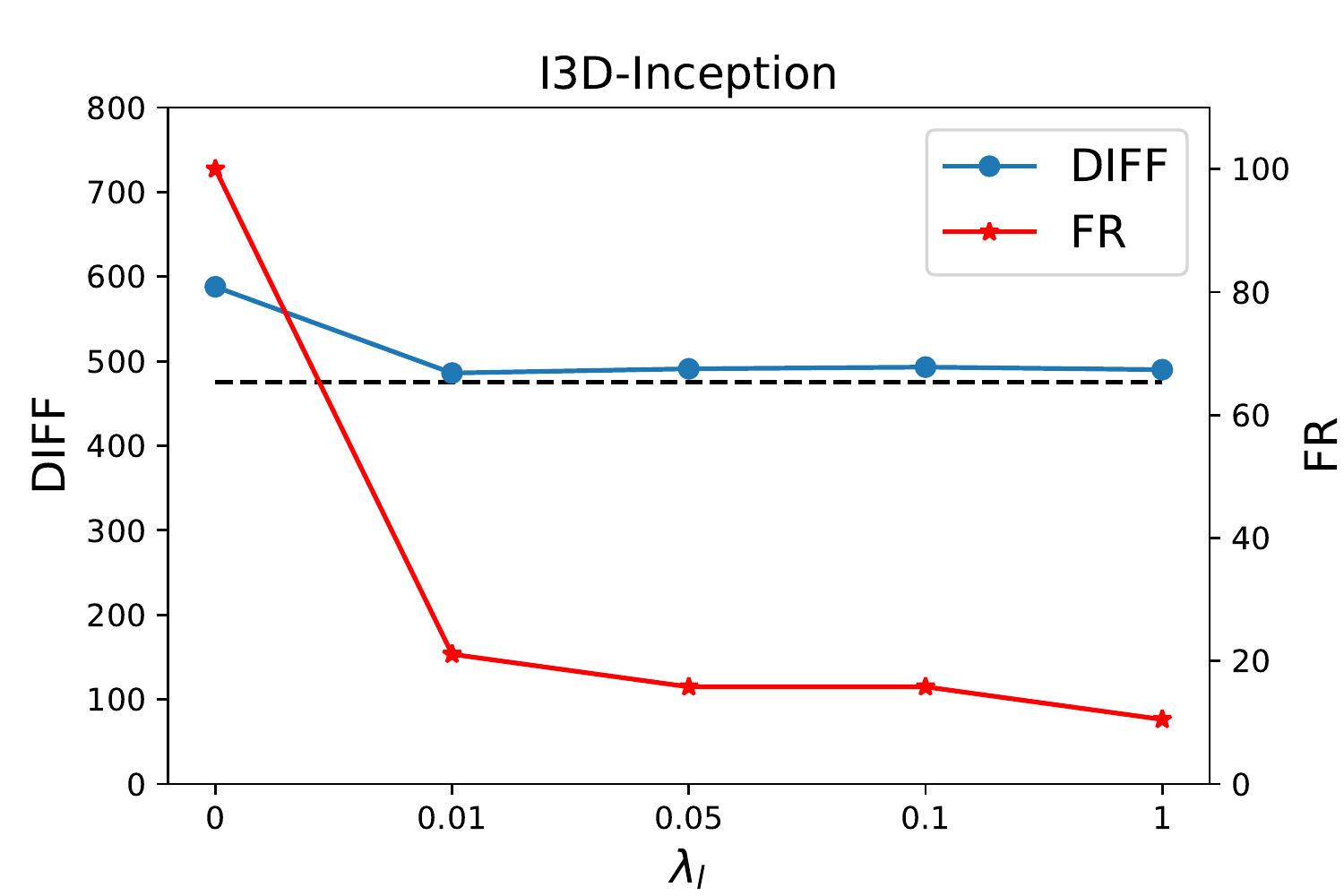}}
        \subfigure{\includegraphics[width=0.45\hsize]{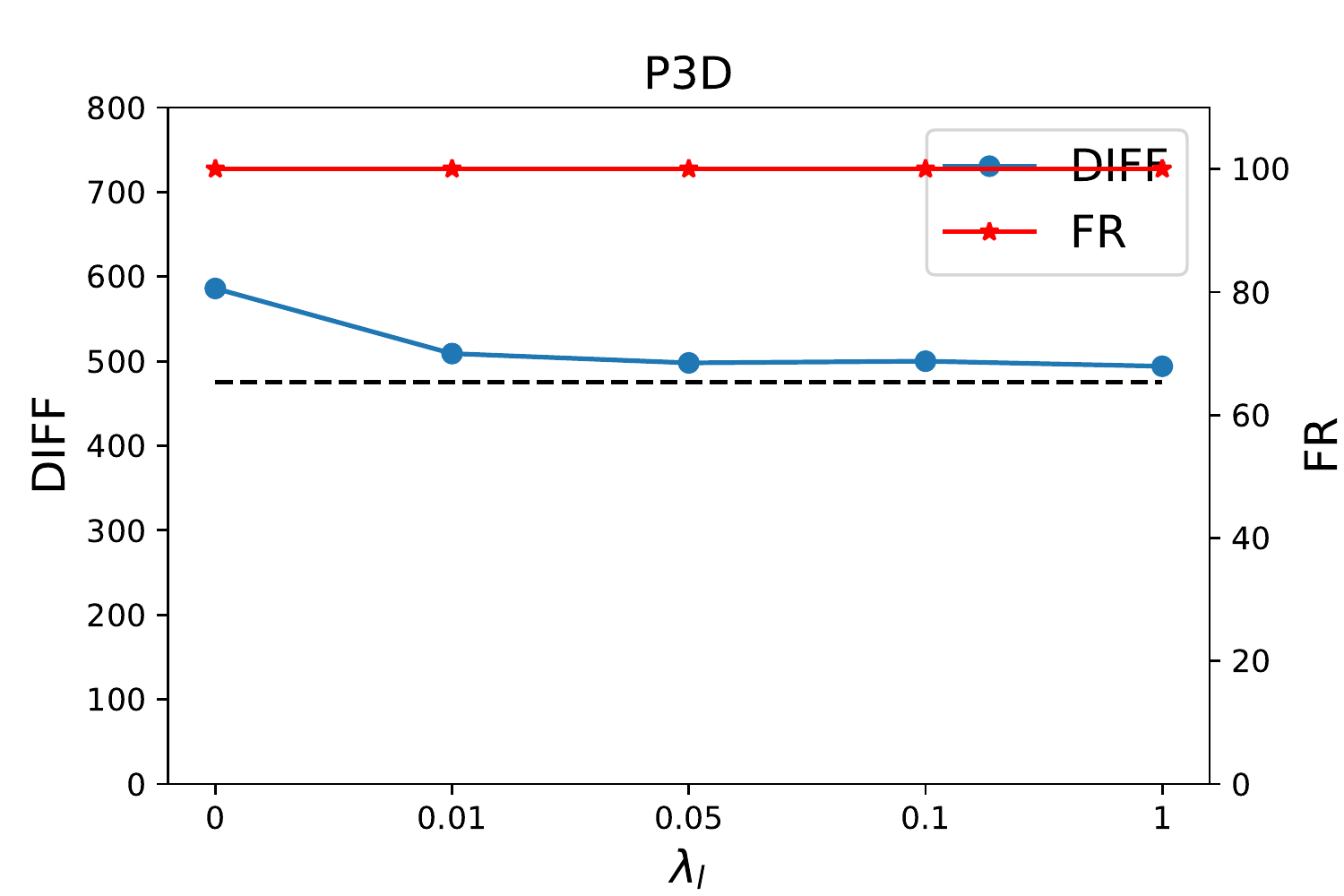}}\\ 
        
           \caption{The attack performance of A\textsuperscript{2}FM-FS with the parameter $\lambda_l$ in different models.}
    \label{figure:DIFF}
\end{figure*}

We validate A\textsuperscript{2}FM-FS with ResNet3D on UCF-101 mentioned in Sec.~4.2. More attack performance in different models under feature similar attack settings is shown in Fig.~\ref{figure:DIFF}.

\end{document}